\documentclass[twoside]{article}

\usepackage{filecontents}
\usepackage[accepted]{aistats2021}
\usepackage[utf8]{inputenc}
\usepackage[english]{babel}
\usepackage{graphicx}
\usepackage{tikz}
\usetikzlibrary{automata,positioning}
\usepackage{natbib}
\usepackage{verbatim}
\usepackage{enumitem}
\usepackage{amsfonts}
\usepackage{algorithm}
\usepackage{float}
\usepackage{units}
\usepackage{subcaption}
\usepackage[noend]{algpseudocode}
\usepackage{booktabs}
\usepackage[normalem]{ulem}
\useunder{\uline}{\ul}{}

\newcommand\ci{\perp\!\!\!\perp}

\newcommand{\G}{{\mathcal G}}
\newcommand{\I}{\mathbb{I}}

\usepackage{amsthm}

\newtheorem{theorem}{Theorem}
\newtheorem{corollary}{Corollary}[theorem]
\theoremstyle{remark}

\theoremstyle{definition}

\usepackage{amsmath}
\usepackage{amssymb}
\usepackage[hidelinks]{hyperref}

\DeclareMathOperator*{\argmin}{argmin}

\newcommand{\trace}[1]{\ensuremath{\text{trace}(#1)}}
\newcommand{\sumall}[1]{\ensuremath{\text{sum}(#1)}}

\newenvironment{thma}[1]{\par\noindent{\bf Theorem #1\ }\em}{\em}

\newenvironment{cora}[1]{\par\noindent{\textbf{Corollary #1} }\em}{\em}

\newcommand{\E}{\mathbb{E}}

\usepackage{multirow}
\usepackage{pifont}
\usepackage{lipsum}
\usepackage{arydshln}

\begin{document}

%

%
\runningauthor{Rohit Bhattacharya, Tushar Nagarajan, Daniel Malinsky, and Ilya Shpitser}

\twocolumn[

\aistatstitle{ Differentiable Causal Discovery Under Unmeasured Confounding }

\aistatsauthor{ Rohit Bhattacharya \\ Johns Hopkins University \\ \texttt{rbhattacharya@jhu.edu} \And Tushar Nagarajan \\ University of Texas at Austin \\ \texttt{tushar@cs.utexas.edu} \AND Daniel Malinsky \\ Columbia University \\ \texttt{d.malinsky@columbia.edu} \vspace{0.65cm} \And  Ilya Shpitser \\ Johns Hopkins University \\ \texttt{ilyas@cs.jhu.edu} \vspace{0.65cm} }
]

\begin{abstract}
	The data drawn from biological, economic, and social systems are often confounded due to the presence of unmeasured variables. 
	Prior work in causal discovery has focused on discrete search procedures for selecting acyclic directed mixed graphs (ADMGs), specifically ancestral ADMGs, that encode ordinary conditional independence constraints among the observed variables of the system. However, confounded systems also exhibit more general equality restrictions that cannot be represented via these graphs, placing a limit on the kinds of structures that can be learned using ancestral ADMGs. In this work, we derive differentiable algebraic constraints that fully characterize the space of ancestral ADMGs, as well as more general classes of ADMGs, arid ADMGs and bow-free ADMGs, that capture all equality restrictions on the observed variables. We use these constraints to cast causal discovery as a continuous optimization problem and design differentiable procedures to find the best fitting ADMG when the data comes from a confounded linear system of equations with correlated errors. We demonstrate the efficacy of our method through simulations and application to a protein expression dataset. Code implementing our methods is open-source and publicly available at \url{https://gitlab.com/rbhatta8/dcd} and will be incorporated into the \href{https://ananke.readthedocs.io/en/latest/index.html}{\texttt{Ananke}} package.
\end{abstract}

\section{INTRODUCTION}

Biological, economic, and social systems are often affected by unmeasured (latent) variables. In such scenarios, statistical and causal models of a directed acyclic graph (DAG) over the observed variables do not faithfully capture the underlying causal process. The most popular graphical structures used to summarize constraints on the observed data distribution are a special class of acyclic directed mixed graphs (ADMGs) with directed and bidirected edges,  known as ancestral ADMGs \citep{richardson2002ancestral}.

Ancestral ADMGs capture all ordinary conditional independence constraints on the observed margin, but they do not capture more general non-parametric equality restrictions, commonly referred to as Verma constraints \citep{verma1990equivalence, tian2002testable, robins1986new}. While ADMGs without the ancestral restriction are capable of capturing all such equality constraints \citep{evans2018margins}, the associated parametric models are not guaranteed to form smooth curved exponential families with globally identifiable parameters -- an important pre-condition for score-based model selection. A smooth parameterization for arbitrary ADMGs is known only when all observed variables are either binary or discrete \citep{evans2014markovian}. For the common scenario when the data comes from a linear Gaussian system of structural equations, the statistical model of an ADMG is almost-everywhere  identified if the ADMG is bow-free \citep{brito2002new}, and is globally identified and forms a smooth curved exponential family if and only if the ADMG is arid \citep{drton2011global, shpitser2018acyclic}. 
From a causal perspective, arid and bow-free ADMGs, like ancestral ADMGs, have the desirable property of preserving ancestral relationships in the underlying latent variable DAG, while also capturing all non-parametric equality restrictions on the observed margin \citep{shpitser2018acyclic}.

We introduce a structure learning procedure for selecting arid, bow-free, or ancestral ADMGs from observational data. Our learning approach is based on reformulating the usual discrete combinatorial search problem into a more tractable constrained continuous optimization program.
Such a reformulation was first proposed by \cite{zheng2018dags} for the special case when the search space is restricted to DAGs. Subsequent extensions such as \cite{yu2019dag}, \cite{zhang2019dvae}, and \cite{zheng2020learning} also restrict the search space in a similar fashion. In this work, we derive differentiable algebraic constraints on the adjacency matrices of the directed and bidirected portions of an ADMG that fully characterize the space of arid ADMGs.
We also derive similar algebraic constraints that characterize the space of ancestral and bow-free ADMGs that are quite useful in practice and connect our work to prior methods. Having derived these differentiable constraints, we select the best fitting graph in the class by optimizing a penalized likelihood-based score. While the constraints we derive in this paper are non-parametric, we focus our causal discovery methods on distributions that arise from linear Gaussian systems of equations. 

Causal discovery methods for learning ancestral ADMGs from data are well developed \citep{spirtes2000causation, colombo2012learning, ogarrio2016hybrid}, but procedures for more general ADMGs are understudied. \cite{hyttinen2014constraint} propose a constraint-based satisfiability solver approach for mixed graphs with cycles. However, their proposal relies on an independence oracle that does not address how to perform valid statistical tests for arbitrarily complex equality restrictions and their procedure may lead to models where the corresponding statistical parameters are not identified (so goodness-of-fit cannot be evaluated). A score-based approach to discovery for linear Gaussian bow-free ADMGs was proposed in \cite{nowzohour2017distributional}. Their method relies on heuristics that may lead to local optima and is not guaranteed to be consistent. Similar issues are faced by the method in \cite{wang2020causal}, which makes a linear non-Gaussian assumption. Currently, there does not exist any consistent fully score-based procedure for learning general ADMGs (besides exhaustive enumeration which is intractable); there are greedy algorithms \citep{bernstein2020ordering} and hybrid greedy algorithms \citep{ogarrio2016hybrid} for ancestral ADMGs, but these are computationally intensive due to the large discrete search space and extending these to arid or bow-free ADMGs would be non-trivial. The procedure we propose has the benefit of being easy to adapt to either ancestral, arid, or bow-free ADMGs while avoiding the need to solve a complicated discrete search problem, instead exploiting state-of-the-art advances in continuous optimization.

Our structure learning procedure for arid and ancestral graphs is consistent in the following sense: asymptotically, convergence to the global optimum implies that the corresponding ADMG is either the true model or one that belongs to the same equivalence class. That is, if the optimization procedure succeeds in finding the global optimum, the resulting graph is either the true underlying structure or one that implies the same set of equality constraints on the observed data. While the $L_0$-regularized objective we propose is non-convex and so our optimization scheme may result in local optima, we show via experiments and application to protein expression data that our proposal works quite well in practice. We believe the algebraic constraints on their own are also valuable for further research at the intersection of non-convex optimization techniques for $L_0$-regularization and causal discovery.

We begin with a motivating example and background on the structure learning problem for partially-observed systems in Sections~\ref{sec:motiv} and~\ref{sec:sems}. In Section~\ref{sec:constraints} we derive differentiable algebraic constraints that characterize arid, bow-free, and ancestral ADMGs. In Section~\ref{sec:methods} we use these to formulate the first (to our knowledge) tractable method for learning arid ADMGs from observational data, by extending the continuous optimization scheme of causal discovery. Simply by modifying the constraint in the optimization program, the same procedure may also be leveraged to learn bow-free or ancestral graphs. Finally we evaluate the performance of our algorithms in simulation experiments and on protein expression data in Section~\ref{sec:experiments}.

\section{MOTIVATING EXAMPLE}
\label{sec:motiv}

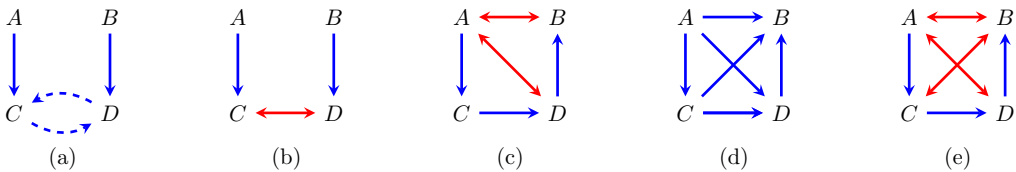
\begin{figure*}[t]
	\begin{center}
		\scalebox{0.85}{
			\begin{tikzpicture}[>=stealth, node distance=1.5cm]
			\tikzstyle{format} = [thick, circle, minimum size=1.0mm, inner sep=0pt]
			\tikzstyle{square} = [draw, thick, minimum size=1.0mm, inner sep=3pt]
			\begin{scope}
			\path[->, very thick]
			node[] (a) {$A$}
			node[right of=a] (b) {$B$}
			node[below of=a] (c) {$C$}
			node[below of=b] (d) {$D$}
			node[below right of=c, xshift=-0.3cm, yshift=0.35cm] (label) {(a)}
			
			(a) edge[blue] (c)
			(b) edge[blue] (d)
			(c) edge[blue, bend right, dashed] (d)
			(d) edge[blue, bend right, dashed] (c)
			;
			\end{scope}
			
			\begin{scope}[xshift=3.5cm]
			\path[->, very thick]
			node[] (a) {$A$}
			node[right of=a] (b) {$B$}
			node[below of=a] (c) {$C$}
			node[below of=b] (d) {$D$}
			node[below right of=c, xshift=-0.3cm, yshift=0.35cm] (label) {(b)}
			
			(a) edge[blue] (c)
			(b) edge[blue] (d)
			(c) edge[red, <->] (d)
			;
			\end{scope}
			
			\begin{scope}[xshift=7cm]
			\path[->, very thick]
			node[] (a) {$A$}
			node[right of=a] (b) {$B$}
			node[below of=a] (c) {$C$}
			node[below of=b] (d) {$D$}
			node[below right of=c, xshift=-0.3cm, yshift=0.35cm] (label) {(c)}
			
			(a) edge[blue] (c)
			(c) edge[blue] (d)
			(d) edge[blue] (b)
			(a) edge[red, <->] (b)
			(a) edge[red, <->] (d)
			;
			\end{scope}
			
			\begin{scope}[xshift=10.5cm]
			\path[->, very thick]
			node[] (a) {$A$}
			node[right of=a] (b) {$B$}
			node[below of=a] (c) {$C$}
			node[below of=b] (d) {$D$}
			node[below right of=c, xshift=-0.3cm, yshift=0.35cm] (label) {(d)}
			
			(a) edge[blue] (c)
			(c) edge[blue] (d)
			(d) edge[blue] (b)
			(a) edge[blue] (b)
			(a) edge[blue] (d)
			(c) edge[blue] (d)
			(c) edge[blue] (b)
			;
			\end{scope}
			
			\begin{scope}[xshift=14cm]
			\path[->, very thick]
			node[] (a) {$A$}
			node[right of=a] (b) {$B$}
			node[below of=a] (c) {$C$}
			node[below of=b] (d) {$D$}
			node[below right of=c, xshift=-0.3cm, yshift=0.35cm] (label) {(e)}
			
			(a) edge[blue] (c)
			(c) edge[blue] (d)
			(d) edge[blue] (b)
			(a) edge[red, <->] (b)
			(a) edge[red, <->] (d)
			(c) edge[red, <->] (b)
			;
			\end{scope}
			
			\end{tikzpicture}
		}
	\end{center}
	\caption{(a) A DAG if $C\rightarrow D$ or $D\rightarrow C$ exists but not both. (b) An ADMG that posits an unmeasured confounder between $C$ and $D$. (c) An (arid) ADMG encoding a Verma constraint between $C$ and $B$. (d) The ancestral version of (c). (e) A non-arid bow-free ADMG that is a super model of (c).}
	\label{fig:examples}
\end{figure*}

To motivate our work, we present an example of how our method may be used to reconstruct complex interactions in a network of genes, which is related to the data application we present in Section~\ref{sec:experiments}.

Consider a scenario in which an analyst has access to gene expression data on four genes: $A, B, C,$ and $D.$ Assume that the analyst is confident (due to prior analysis or background knowledge) about the structure corresponding to non-dashed edges shown in Fig.~\ref{fig:examples}(a), i.e., that $A$ regulates $C$ and $B$ regulates $D$ but $A$ and $B$ are independent. This leaves an important ambiguity regarding regulatory explanations of co-expression of genes $C$ and $D$.

An observed correlation between $C$ and $D$ 
may be explained in different ways that provide very different mechanistic interpretations. If the hypothesis class is restricted to DAGs, the only explanations available to the analyst are that $C$ is a cause of $D$ or vice-versa as shown in Fig.~\ref{fig:examples}(a). If the analyst proceeds with either of these explanations and performs a gene-knockout experiment where $C$ (or $D$) is removed but sees no change in $D$ (respectively $C$), then the causal DAG fails to be a faithful representation of the true underlying mechanism.
The correlation may instead be explained by an ADMG as in Fig.~\ref{fig:examples}(b) where $C \leftrightarrow D$ indicates that $C$ and $D$ are dependent due to the presence of at least one unmeasured confounding gene that regulates both of them. That is, if we had data on these unmeasured genes $U$ the corresponding DAG would have contained a structure $C \leftarrow U \rightarrow D.$ However, given observations only on $A,B,C,D$, Fig.~\ref{fig:examples}(b) provides a faithful representation of this underlying mechanism on the observed variables. It correctly encodes that intervention on $C$ or $D$ has no downstream effects on the other.

Importantly, each of these different explanations are not just different from a mechanistic point of view but also imply different independence restrictions on the observed data. The two DAGs in Fig.~\ref{fig:examples}(a) imply that $A \ci D \mid C$ or  $B \ci C \mid D$ respectively, whereas Fig.~\ref{fig:examples}(b) implies $A \ci D$ and $B \ci C$.
Hence, a causal discovery procedure that seeks the best fitting structure from the hypothesis class of ADMGs, will be able to distinguish between these different explanations and choose the correct one.

Some mechanisms, such as the one shown in Fig.~\ref{fig:examples}(c), are not distinguishable using ordinary conditional independence statements alone. In this graph, the only pair of genes with no edge between them is $B$ and $C$. The absence of this edge implies that $C$ does not directly regulate the expression of $B$ and only does so through $D$. This missing edge does not correspond to any ordinary conditional independence (there are no independence constraints implied by the model at all), but does encode a Verma constraint, namely that $B \ci C \mid D$ in a re-weighed distribution derived from the joint, $p(A,B,C,D)/p(C | A)$. 

The following ADMG classes will be important in this work.
An ADMG $\G = (V,E)$ is said to be \emph{ancestral} if for any pair of vertices $V_i, V_j \in V$, a directed path $V_i \rightarrow \cdots \rightarrow V_j$ and bidirected edge $V_i \leftrightarrow V_j$ do not both appear in $\G$.
An ADMG $\G$ is said to be \emph{arid} if it does not contain any \emph{c-trees}. A c-tree is a subgraph of $\G$ whose directed edges form an arborescence (the directed graph analogue of a tree) and bidirected edges form a single bidirected connected component within the subgraph. 
It is easy to confirm that the ADMG in Fig.~\ref{fig:examples}(b) is ancestral while the one in Fig.~\ref{fig:examples}(c) is arid but not ancestral. An ADMG is called \emph{bow-free} if for any pair of vertices, $V_i \rightarrow V_j$ and $V_i \leftrightarrow V_j$ do not both appear in $\G.$ A graph that is bow-free but neither arid nor ancestral is displayed in Fig.~\ref{fig:examples}(e). The relation between these graph classes is the following: 
\begin{align*}
	\text{Ancestral} \subset \text{Arid} \subset \text{Bow-free}
\end{align*}
Ancestral graphs can ``hide'' certain important information because they encode only ordinary conditional independence constraints. An ancestral graph that encodes the same ordinary independence constraints as the arid graph in Fig.~\ref{fig:examples}(c) is shown in Fig.~\ref{fig:examples}(d).
It is a complete graph since there are no conditional independence constraints in Fig.~\ref{fig:examples}(c). That is, the absence of any $C \rightarrow B$ edge in Fig.~\ref{fig:examples}(c) is ``masked'' to preserve the ancestrality property. We can potentially learn a more informative structure if we do not limit our search space to the class of ancestral graphs.


\section{GRAPHICAL INTERPRETATION OF LINEAR SEMs}
\label{sec:sems}

In this section, we review linear SEMs and their graphical representations. 
We use capital letters (e.g. $V$) to denote sets of variables and nodes on a graph interchangeably and capital letters with an index (e.g. $V_i$) to refer to a specific variable or node in $V$. We also make use of the following standard matrix notation: $A_{ij}$ refers to the element in the $i^{th}$ row and $j^{th}$ column of a matrix $A,$ indexing $A_{-i,-j}$ refers to the sub matrix obtained by excluding the $i^{th}$ row and $j^{th}$ column of $A,$ and $A_{:, i}$ refers to the $i^{th}$ column of $A.$

\subsection{Linear SEMs and DAGs}

Consider a linear SEM on $d$ variables parameterized by a weight matrix $\theta \in \mathbb{R}^{d \times d}.$ For each variable $V_i \in V,$ we have a structural equation $V_i \gets \sum_{V_j \in V} \theta_{ji} V_j  + \epsilon_i,$
%
%
where the noise terms $\epsilon_i$ are mutually independent. That is, $\epsilon_i \ci \epsilon_j$ for all $i \not=j.$ Let $\G(\theta)$ and $D(\theta) \in \{0, 1\}^{d \times d}$ be the induced directed graph and corresponding binary adjacency matrix obtained as follows: $V_i \rightarrow V_j$ exists in $\G(\theta)$ and $D(\theta)_{ij} = 1$ if and only if $\theta_{ij} \not=0$. The induced graph $\G$ has no directed cycles if and only if $\theta$ 
can be made upper-triangular via a permutation of vertex labelings
\citep{mckay2004acyclic}. Such an SEM is said to be \emph{recursive} or \emph{acyclic} and the corresponding probability distribution $p(V)$ is said to be Markov with respect to the DAG $\G(\theta).$ This means that conditional independence statements in $p(V)$ can be read off from $\G$ via the well-known d-separation criterion \citep{pearl2009causality}.

\subsection{Systems with Unmeasured Confounding}

A set of observed variables is called \emph{causally insufficient} if there exist unobserved variables, commonly referred to as latent confounders, that cause two or more observed variables in the system. 
In the linear SEM setting, unmeasured variables manifest as correlated errors \citep{pearl2009causality}. 
Such an SEM on $d$ variables can be parameterized by two real-valued matrices $\delta, \beta \in \mathbb{R}^{d\times d}$ as follows. For each $V_i \in V,$ we have a structural equation $V_i \gets \sum_{V_j \in V} \delta_{ji} V_j  + \epsilon_i,$
%
%
and the dependence between the noise terms $\epsilon = (\epsilon_1,...,\epsilon_d)$ is summarized via their covariance matrix $\beta = \E[\epsilon \epsilon^T]$. In the case when each noise term $\epsilon_i$ is normally distributed the induced distribution $p(V)$ is jointly normal with mean zero and covariance matrix $\Sigma = (I - \delta)^{-T} \beta (I - \delta)^{-1}.$
The induced graph $\G$ is a mixed graph consisting of directed ($\rightarrow$) and bidirected ($\leftrightarrow$) edges and can be represented via two adjacency matrices $D$ and $B$. $V_i \rightarrow V_j$ exists in $\G$ and $D_{ij}=1$ if and only if $\delta_{ij} \not= 0.$ $V_i \leftrightarrow V_j$ exists in $\G$ and $B_{ij} = B_{ji}=1$ if and only if $\beta_{ij} \not= 0.$ That is, the adjacency matrix $B$ corresponding to bidirected edges in $\G$ is symmetric as the covariance matrix $\beta$ itself is symmetric (and positive definite).

We consider three classes of mixed graphs to represent causally insufficient linear SEMs: 
ancestral, arid, and bow-free ADMGs. All of these have no directed cycles and lack specific substructures as defined in the previous section. A distribution $p(V)$ induced by a linear Gaussian SEM is said to be Markov with respect to an ADMG $\G$ if absence of an edge between $V_i$ and $V_j$ implies $\delta_{ij}=\delta_{ji}=\beta_{ij}=\beta_{ji}=0$ which in turn implies equality restrictions on the support of all possible covariance matrices $\Sigma(\G)$ by forcing certain polynomial functions of entries in the covariance matrix to evaluate to 0 \citep{yao2019constraints}. To facilitate causal discovery, we assume a generalized version of faithfulness, similar to the one  in \cite{ghassami2020characterizing}, stating that if a distribution $p(V)$ is induced by a linear Gaussian SEM where $\delta_{ij}=\delta_{ji}=\beta_{ij}=\beta_{ji}=0$ then there is no edge present between $V_i$ and $V_j$ in $\G.$ In other words, we define $p(V)$ to be Markov and faithful with respect to $\G$ if absence of edges in $\G$ occurs if and only if the corresponding entries in $\delta$ and $\beta$ are $0.$


As a concrete example, let $\Sigma$ denote the covariance matrix of standardized normal  random variables $A, B, C, D$ drawn from a linear SEM that is Markov with respect to the ADMG in Fig.~\ref{fig:examples}(c), and let $\delta$ and $\beta$ denote the corresponding normalized coefficient matrices. By standard rules of path analysis \citep{wright1921correlation, wright1934method}, the Verma constraint due to the missing edge in Fig.~\ref{fig:examples}(c)
corresponds to the equality constraint: 
\begin{align*}
	\Sigma_{BC} - \delta_{CD}\delta_{DB} - \delta_{AC}\beta_{AB} - \delta_{AC}\beta_{AD}\delta_{DB} = 0.
\end{align*}
Since entries in the covariance matrix are rational functions of $\delta$ and $\beta,$ the above constraint can be re-expressed solely in terms of entries in $\Sigma.$ Our faithfulness assumption is used to ensure that such polynomial functions of the covariance matrix 
do not ``accidentally'' evaluate to zero, and only do so due to a missing edge in the underlying ADMG.

As mentioned earlier, ancestral ADMGs cannot encode such generalized equality restrictions but arid and bow-free ADMGs can. For any ADMG $\G,$ an arid ADMG that shares all non-parametric equality constraints with $\G$ may be constructed by an operation called maximal arid projection \citep{shpitser2018acyclic}. 
We also consider bow-free ADMGs because the algebraic constraint characterizing the bow-free property is simpler than the one characterizing the arid property. Though the lack of global identifiability in bow-free ADMG models (only almost everywhere identifiable) can pose problems for model convergence, we confirm in our experiments that enforcing only the weaker bow-free property is often sufficient for accurate causal discovery in practice.

\section{DIFFERENTIABLE ALGEBRAIC CONSTRAINTS}
\label{sec:constraints}

We now introduce differentiable algebraic constraints that precisely characterize when the parameters of a linear SEM induce a graph that belongs to any one of the ADMG classes described in the previous section. Our results are summarized in Table~\ref{tab:constraints} in terms of the binary adjacency matrices but as we explain below, the results extend in a straightforward manner to real-valued matrices that parameterize a linear SEM. In Table~\ref{tab:constraints}, $A \circ B$ denotes the Hadamard (elementwise) matrix product between $A$ and $B$ and $e^A$ denotes the exponential of a square matrix $A$ defined as the infinite Taylor series, $e^A = \sum_{k=0}^{\infty} \frac{1}{k!} A^k$. We formalize the properties of our constraints in the following theorem.

\begin{theorem}
	\label{thm:constraints}
	The constraints shown in Table~\ref{tab:constraints} are satisfied
	if and only if the adjacency matrices satisfy the relevant property of ancestrality, aridity, and bow-freeness respectively.
\end{theorem}

We defer formal proofs to the Appendix but briefly provide intuition for our results. For a binary square matrix $A,$ corresponding to a directed/bidirected adjacency matrix, the entry $A_{ij}^k$ counts the number of directed/bidirected walks of length $k$ from $V_i$ to $V_j;$ see for example \cite{butler2008eigenvalues}. For $k=0,$ $D^k$ is the identity matrix by definition and for $k \geq 1,$ each diagonal entry of the matrix $D^k$ appearing in the infinite series $e^D$  thus corresponds to the number of directed walks of length $k$ from a vertex back to itself, i.e., the number of directed cycles of length $k$. The quantity $\trace{e^D}-d$ is therefore a weighted count of the number of directed cycles in the induced graph and is zero precisely when no such cycles exist. Hence, this term appears in all algebraic constraints presented in Table~\ref{tab:constraints} as requiring $\trace{e^D}-d=0$ enforces acylicity. 

\begin{algorithm}[t]
	\caption{\textproc{Greenery} $(D, B)$} \label{alg:greenery}
	\begin{algorithmic}[1]
		\State $\text{greenery} \gets 0\  \text{ and }\   I \gets d\times d \text{ identity matrix} $
		\For {$i \text{ in } (1, \dots, d)$}
		\State $D_f, B_f \gets D, B$
		\For {$ j \text{ in } (1, \dots, d-1)$}
		\State $t \gets \text{row sums of } e^{B_f} \circ D_f$ \Comment{$1\times d$ vector}
		\State $f \gets \tanh(t + I_i)$ \label{alg:greenery_fix} \Comment{$1\times d$ vector}
		\State $F \gets [f^T; \dots; f^T]^T$\Comment{$d\times d$ matrix}
		\State $D_f\gets D_f \circ F \text{ and }   B_f \gets B_f \circ F \circ F^T$ \label{alg:greenery_fix_end}
		\EndFor
		\State $C \gets e^{D_f}\circ e^{B_f}$
		\State greenery $+= \text{sum}(C_{:,i})$ \Comment{sum of $i^{th}$ column}
		\EndFor
		\State \textbf{return} $\text{greenery} - d$
	\end{algorithmic}
\end{algorithm}
\renewcommand{\arraystretch}{1.5}
\begin{table}
	\begin{center}
		\begin{tabular}{ |c|c| } 
			\hline
			\textbf{ADMG} &  \textbf{Algebraic Constraint} \\
			\hline
			Ancestral  & $\trace{e^D} - d + \sumall{e^D \circ B} = 0$ \\
			Arid &  $\trace{e^D} - d + \textproc{Greenery}(D,B) = 0$ \\
			Bow-free &  $\trace{e^D} - d + \sumall{D\circ B} = 0$ \\
			\hline	
		\end{tabular}
		\caption{Differentiable algebraic constraints that characterize the space of binary adjacency matrices that fall within each ADMG class. The \textproc{Greenery} algorithm to penalize c-trees is described in Algorithm~\ref{alg:greenery}. }
		\label{tab:constraints}
	\end{center}
\end{table}
Similar reasoning can be used to show that requiring $\text{sum}(e^D \circ B)=0$ enforces ancestrality. An entry $i, j$ of the matrix $D^k \circ B$ appearing in the infinite series counts the number of violations of ancestrality due to a directed path from $V_i$ to $V_j$ of length $k$ and a bidirected edge $V_i \leftrightarrow V_j.$ The sum of all such terms is then precisely zero when the induced graph is ancestral. The bow-free constraint $\text{sum}(D \circ B)=0$ is simply a special case of the ancestral constraint where directed paths of length $\geq 2$ need not be considered.

C-trees are known to be linked to the identification of causal parameters, specifically, the effect of each variable's parents on the variable itself \citep{shpitser2006identification, huang2006pearl}. 
The outer loop of Algorithm~\ref{alg:greenery} iterates over each vertex $V_i$ to determine if there is a $V_i$-rooted c-tree. The inner loop performs the following recursive simplification at most $d-1$ times. At each step, the sum of the $j^{th}$ row of the matrix  $e^{B_f} \circ D_f$ is zero if and only if there are no bidirected paths from $V_j$ to any of its direct children. If this criterion -- called primal fixability -- is met, the effect of $V_j$ on its children is identified and the post-intervention distribution can be summarized by a new graph with all incoming edges into $V_j$ removed \citep{bhattacharya2020semiparametric}. Lines~\ref{alg:greenery_fix}-\ref{alg:greenery_fix_end} are the algebraic operations that correspond to deletion of incoming directed and bidirected edges into primal fixable vertices, except $V_i$ itself as it is the root node of interest. The hyperbolic tangent function is used to ensure that recursive applications of the operation do not result in large values. At the end of the recursion, the co-existence of directed and bidirected paths to $V_i$ imply the existence of a c-tree. Hence, the quantity $\text{sum}(C_{:, i})$ is non-negative and is zero if and only if there is no $V_i$-rooted c-tree. Concrete examples of applying Algorithm~\ref{alg:greenery}, and its connections to primal fixing are provided in Appendix~\ref{app:greenery}.

It is easy to see that the above results and intuitions can be applied to arbitrary non-negative real-valued matrices $D$ and $B.$ Theorem~\ref{thm:constraints} then extends in a straightforward manner to parameters of a linear SEM by noting that for any real-valued matrix $A,$ the matrix $A \circ A$ is real-valued and non-negative.

\begin{corollary}
	\label{cor:constraints_sem}
	The result in Theorem~\ref{thm:constraints} and the constraints in Table~\ref{tab:constraints} can be applied to linear SEMs by plugging in $D \equiv \delta \circ \delta$ and $B \equiv \beta' \circ \beta',$ where $\beta_{ij}'=\beta_{ij}$ for $i\not=j$ and $0$ otherwise.
\end{corollary}

Finally, while the matrix exponential makes theoretical arguments simple, the resulting constraints are not numerically stable as pointed out in \cite{yu2019dag}. The following corollary provides a more stable alternative that we use in our implementations.
\begin{corollary}
	\label{cor:constraints_power}
	The results in Theorem~\ref{thm:constraints} and Corollary~\ref{cor:constraints_sem} hold if every occurrence of a matrix exponential $e^{A}$ is replaced with the matrix power $(I + c A)^d$ for any $c > 0,$ where $I$ is the identity matrix.
\end{corollary}


\section{DIFFERENTIABLE SCORE BASED CAUSAL DISCOVERY}
\label{sec:methods}

Let $\theta$ be the parameters of a linear SEM. We use $\theta$ here to refer to a generic parameter vector that can be reshaped into the appropriate parameter matrices $\delta,$ and $\beta$ as discussed in Section~\ref{sec:sems}. Let $\G(\theta)$ be the corresponding induced graph. Given a dataset $X\in \mathbb{R}^{n\times d}$  drawn from the linear SEM and a hypothesis class $\mathbb{G}$ that corresponds to one of ancestral, arid, or bow-free ADMGs, the combinatorial problem  of finding an optimal set of parameters $\theta^* \in \Theta$ that minimizes some score $f({X}; \theta)$ such that $\G(\theta) \in \mathbb{G}$ can be rephrased as a more tractable continous program.
\begin{equation}
	\!
	\begin{aligned}[c]
		\min_{\theta \in \Theta}\  f({X}; \theta) \\
		\text{s.t. }\  \G(\theta) \in \mathbb{G} 
	\end{aligned}
	\hspace{0.75cm} \Longleftrightarrow \hspace{0.75cm}
	\begin{aligned}[c]
		\min_{\theta \in \Theta}\  f({X}; \theta) \\
		\text{s.t. }\  h(\theta) = 0.
	\end{aligned}
	\label{eq:reformulation}
\end{equation}

The results in the previous section in Theorem~\ref{thm:constraints}, its Corollaries and Table~\ref{tab:constraints} tell us how to pick the appropriate function $h(\theta)$ for each hypothesis class $\mathbb{G}.$ We now discuss choices of score function $f({X}; \theta)$ and procedures to minimize it for different hypothesis classes.

\subsection{Choice of Score Function}

Given a dataset ${X} \in \mathbb{R}^{n\times d},$ the Bayesian Information Criterion (BIC) is given by $-2\ln({\cal L}({X}; \theta)) + \ln(n)\sum_{i=1}^{\text{dim}(\theta)} \I(\theta_i \not=0),$ where ${\cal L(\cdot)}$ is the likelihood function and $\text{dim}(\theta)$ is the dimensionality of $\theta.$
The BIC is consistent for model selection in curved exponential families \citep{schwarz1978estimating, haughton1988on}, i.e., as  $n\to\infty$ the BIC attains its minimum at the true model (or one that is observationally equivalent to it). This results in the following desirable theoretical property when the BIC is used as our objective function. 

\begin{theorem}
	Let $p(V; \theta^*)$ be a distribution in the curved exponential family that is Markov and faithful  with respect to an arid ADMG $\G^*$. Finding the global optimum of the continuous program in display~(\ref{eq:reformulation}) with $f\equiv BIC$ yields an ADMG $\G(\theta)$ that implies the same equality restrictions as $\G^*.$
	\label{lem:optimum}
\end{theorem}

However, the presence of the indicator function makes the BIC non-differentiable and optimization of $L_0$ objectives like the BIC is known to be NP-hard \citep{natarajan1995sparse}. While $L_1$ regularization is a popular alternative, it often leads to inconsistent model selection and overshrinkage of coefficients \citep{fan2001variable}.
Several procedures have been devised in order to provide approximations of the BIC score; see \cite{huang2018constructive} for an overview. In this work, we consider the approximate BIC (ABIC) obtained via replacement of the indicator function with the hyperbolic tangent function as outlined in \cite{su2016sparse} and \cite{nabi2017coxphmic}. That is, we seek to optimize $-2\ln({\cal L}({X}; \theta)) + \lambda \sum_{i=1}^{\text{dim}(\theta)} \tanh(c|\theta_i|),$ where $c>0$ is a constant that controls the sharpness of the approximation of the indicator function and $\lambda$ controls the strength of regularization. As highlighted in \cite{su2016sparse}, the ABIC is relatively insensitive to the choice of $c.$ The main hyperparameter is the regularization strength $\lambda.$ In our experiments we set $c=\ln(n)$ and report results for different choices of $\lambda.$ In the next section we discuss our strategy to optimize the ABIC subject to the constraint that $\theta$ induces a valid ADMG within a hypothesis class $\mathbb{G}.$

\subsection{Solving the Continuous Program}

\begin{algorithm}[t]
	\caption{ \textproc{Regularized RICF} } \label{alg:pseudo-ricf}
	\begin{algorithmic}[1]
		\State \textbf{Inputs}: $(X, \text{tol}, \text{max iterations}, h, \rho, \alpha, \lambda)$
		
		\State Initialize estimates $\delta^t$ and $\beta^t$ and set $c=\ln(n)$
		\State Define $\text{LS}(\theta)$ as $\frac{1}{2n}\sum_{i=1}^d||X_{:, i} - X \delta_{:, i} - Z^{(i)}\beta_{:, i} ||_2^2$
		\For{$t \text{ in } (1, \dots, \text{max iterations})$}
		
		\State $\forall i \in (1, \dots, d)$ compute $\epsilon_i \gets X_{:,i} - \delta^t_{:,i} X$
		
		\State $\forall i \in (1, \dots, d)$ compute $Z^{(i)} \in \mathbb{R}^{n \times d}$ as
		
		\hspace{0.7cm}$Z^{(i)}_{:, i}=0$ and $Z^{(i)}_{:, -i} \gets \epsilon_{-i} {(\beta_{-i, -i}^t})^{-T}$
		\State $\delta^{t+1}, \beta^{t+1} \gets \argmin_{\theta \in \Theta} \big\{ \text{LS}(\theta) + \frac{\rho}{2} |h(\theta)|^2 $
		
		\hspace{1.8cm}$ + \ \alpha h(\theta) + \lambda \sum_{i=1}^{\text{dim}(\theta)}\tanh(c|\theta_i|) \big\}$ \label{line:ricf_primal}
		\State $\forall i \in (1, \dots, d)$ compute $\epsilon_i \gets X_{:,i} - \delta^{t+1}_{:,i} X$ 
		\State $\forall i \in (1, \dots, d)$ set $\beta^{t+1}_{ii} \gets \text{var}(\epsilon_i)$ 
		\If{$||\delta^{t+1} - \delta^{t}  + \beta^{t+1} - \beta^{t}|| < \text{tol}$} break
		\EndIf
		\EndFor
		
		\State \textbf{return} $\delta^t, \beta^t$
	\end{algorithmic}
\end{algorithm}

\begin{algorithm}[t]
	\caption{\textproc{Differentiable Discovery} }
	\label{alg:discovery}
	\begin{algorithmic}[1]
		\State \textbf{Inputs}: $(X, \text{tol}, \text{max iterations}, s, h, \lambda, r \in (0, 1))$
		
		\State Initialize $\theta^t, \alpha^t, m^t \gets 1$ 
		\While{$t  < \text{max iterations} \text{ and } h(\theta^t) > \text{tol}$}
		\State $\theta^{t+1} \gets$ $\theta^*$ from \textproc{Regularized RICF} with
		
		\hspace{1cm}inputs $(X, 10^{-4}, m^t, h, \rho, \alpha^t, \lambda)$  
		
		\hspace{1cm}where $\rho$ is such that  $h(\theta^{*}) < rh(\theta^t)$ \label{line:rho_rule}
		\State $\alpha^{t+1}\gets \alpha^t + \rho h(\theta^{t+1}) \text{ and }  m^{t+1} \gets m^t + s$
		\EndWhile
		\State \textbf{return} $\G(\theta^t)$
	\end{algorithmic}
\end{algorithm}

We formulate the optimization objective as minimizing the ABIC subject to one of the algebraic equality constraints in Table~\ref{tab:constraints}. We use the augmented Lagrangian formulation~\citep{bertsekas1997nonlinear} to convert the problem into an unconstrained optimization problem with a quadratic penalty term, which can be solved using a dual ascent approach. Specifically, in each iteration we first solve the primal equation:
\begin{align*}
	\min_{\theta \in \Theta} \ \text{ABIC}_\lambda(X;\theta) + \frac{\rho}{2}|h(\theta)|^2 + \alpha h(\theta),
\end{align*}
where $\rho$ is the penalty weight and $\alpha$ is the Lagrange multiplier. Then we solve the dual equation $\alpha \leftarrow \alpha + \rho h(\theta^*).$ Intuitively, optimizing the primal objective with a large value of $\rho$ would force $h(\theta)$ to be very close to zero thus satisfying the equality constraint.
%
%

However, unlike DAG models, maximum likelihood estimation of parameters under the restrictions of an ADMG does not correspond to a simple least squares regression that can be solved in one step. 
\cite{drton2009computing} proposed an iterative procedure known as Residual Iterative Conditional Fitting (RICF) that produces a sequence of maximum likelihood estimates for $\delta$ and $\beta$ under the constraints implied by a fixed ADMG $\G.$ Each RICF step is guaranteed to produce better estimates than the previous step and the overall procedure is guaranteed to converge to a local optimum or saddle point when $\G(\theta)$ is arid/ancestral, i.e., globally identified \citep{drton2011global}.

In Algorithm~\ref{alg:pseudo-ricf} we describe a modification of RICF that directly inherits the aforementioned properties with respect to the regularized maximum likelihood objective, and can be used to solve the primal equation of our procedure.
Briefly, for Gaussian ADMG models, maximization of the likelihood corresponds to minimization of a least squares regression problem where each variable $i$ is regressed on its direct parents $V_j \rightarrow V_i$ and pseudo-variables $Z$ formed from the residual noise terms and bidirected coefficients of its siblings $V_j \leftrightarrow V_i.$ At each RICF step, we compute $Z$ with respect to the current parameter estimates, and then solve the primal equation in line~\ref{line:ricf_primal} of the algorithm. We repeat this until convergence or a pre-specified maximum number of iterations. As RICF is not expected to converge during initial iterations of the augmented Lagrangian procedure when the penalty applied to $h(\theta)$ is quite small (resulting in non-arid graphs), we start with a small number of maximum RICF iterations and at each dual step increment this number. The penalty $\rho$ applied to $h(\theta)$ is increased according to a fixed schedule where $\rho$ is multiplied by a factor of 10 (up to a maximum value of $10^{16}$) each time the inequality in line~\ref{line:rho_rule} of the algorithm is not satisfied. Our simulations show this works quite well in practice with convergence of the algorithm obtained typically within 10-15 steps of the augmented Lagrangian procedure.

We summarize our structure learning algorithm in Algorithm~\ref{alg:discovery}. 
Though optimization of the objective in display~(\ref{eq:reformulation}) is non-convex, standard properties of dual ascent procedures as well as the RICF algorithm guarantee that at each step in the process we recover parameter estimates that do not increase the objective  we are trying to minimize. Further, per Theorem~\ref{lem:optimum}, if optimization of the ABIC objective for a given level of $\lambda$ provides a good enough approximation of the BIC, the global minimizer (if found by our optimization procedure) yields a graph that implies the same equality restrictions as the true graph. 

\subsection{Reporting Equivalent Structures}
Our procedure only reports a single ADMG but there may exist multiple ADMGs that imply the same equality restrictions on the observed data. In the linear Gaussian setting, exact recovery of the skeleton of the ADMG (i.e., adjacencies without any orientations) is possible, but complete determination of all edge orientations is not. Reporting the uncertainty in edge orientations is important for downstream causal inference tasks. When limiting our hypothesis class $\mathbb{G}$ to ancestral ADMGs, the non-parametric equivalence class can be represented via a Partial Ancestral Graph (PAG). After obtaining a single ADMG using our procedure, we can easily reconstruct its equivalence class using rules in \cite{zhang2008completeness} to create the summary PAG. For arid and bow-free ADMGs, a full theory of equivalence that captures Verma constraints is still an open problem. Thus, while we are able to recover the exact skeleton, we coarsen reporting of edge orientations by converting the estimated ADMG into an ancestral ADMG and reporting the PAG. Connections in this PAG may be pruned using sound rules from \cite{nowzohour2017distributional} and \cite{zhang2020simultaneous} though we do not pursue this approach here.  Deriving a summary structure that captures the class of all ADMGs that are equivalent up to equality restrictions is an important problem but outside the scope of this work.

\section{EXPERIMENTS}
\label{sec:experiments}


\begin{figure*}[t]
	\centering
	\includegraphics[scale=0.35]{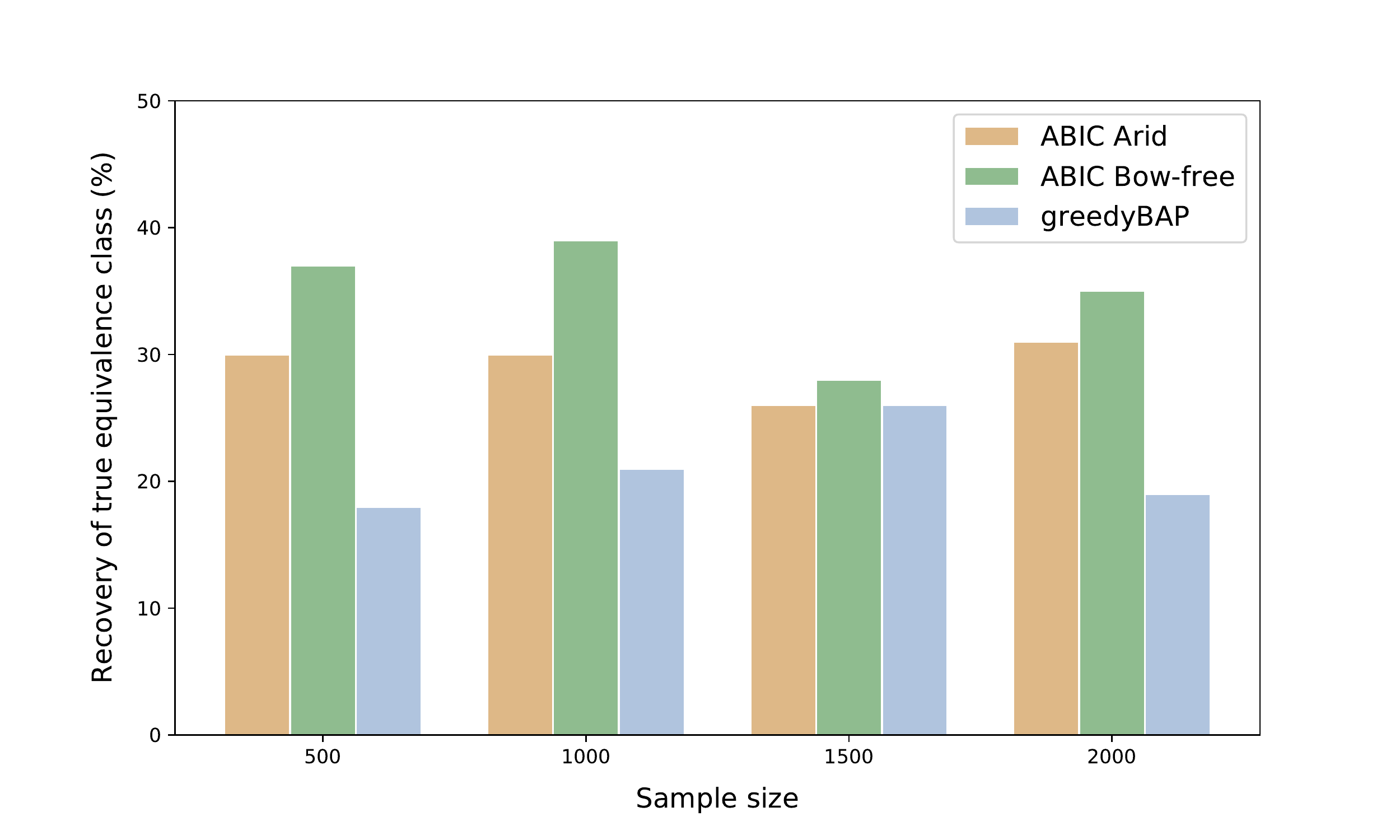}
	\includegraphics[scale=0.35]{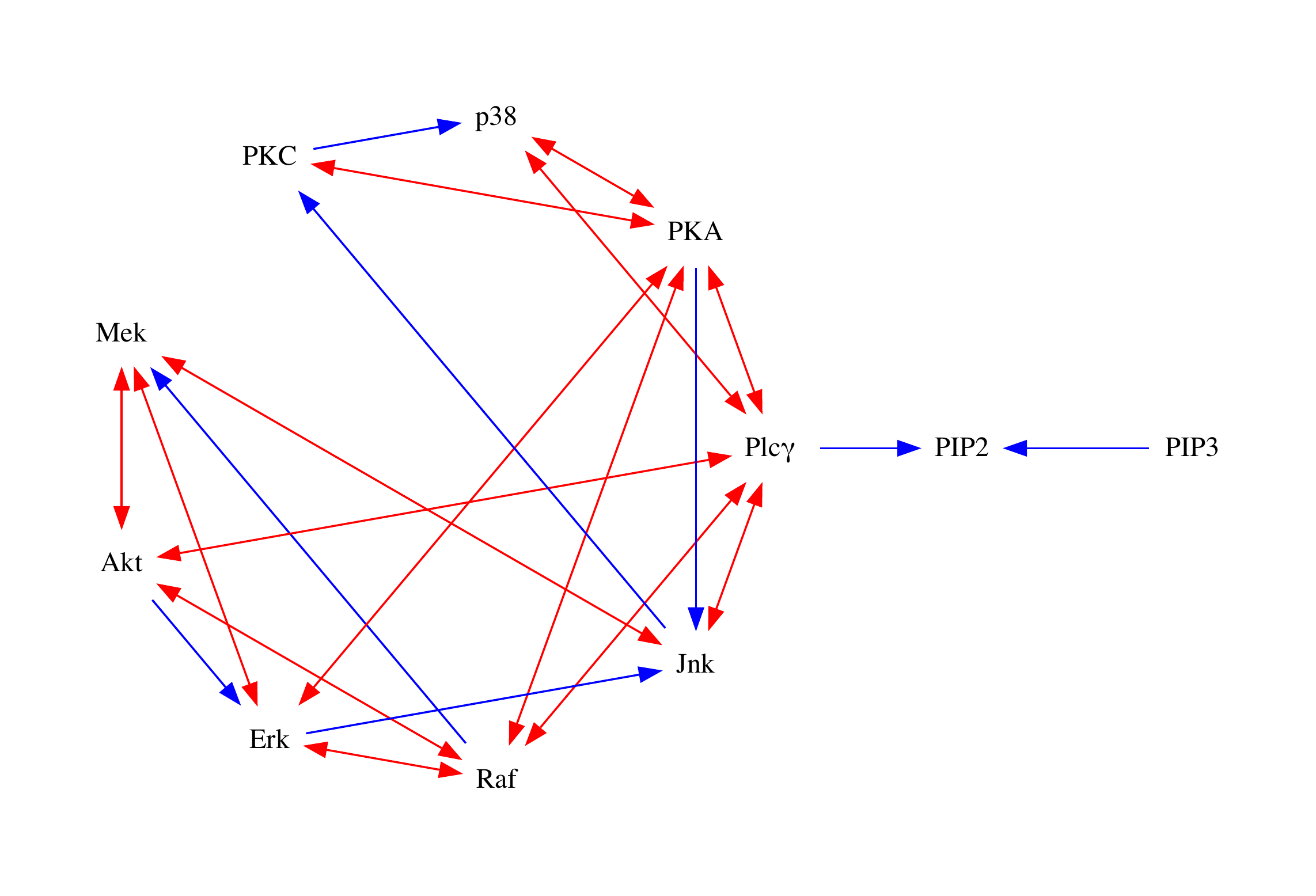}
	\caption{Left: Rate of recovery of the true equivalence class of an ADMG with a Verma constraint as a function of sample size. Right: Application of the ABIC bow-free method to the \cite{sachs2005causal} dataset.}
	\label{fig:experiments}
\end{figure*}

\begin{table*}[t]
	\resizebox{\textwidth}{!}{
		
		\begin{tabular}{l|ll|ll|ll|}
			\cline{2-7}
			& \multicolumn{2}{c|}{\textsc{Skeleton}}  & \multicolumn{2}{c|}{\textsc{Arrowhead}} & \multicolumn{2}{c|}{\textsc{Tail}}    \\ \hline
			\multicolumn{1}{|l|}{Method}        & tpr $\uparrow$ & fdr $\downarrow$ & tpr $\uparrow$& fdr$\downarrow$ & tpr $\uparrow$& fdr $\downarrow$ \\ \hline
			\multicolumn{1}{|l|}{gBAP~\citep{nowzohour2017distributional}} 
			& 0.80       & 0.30       & 0.41       & 0.58       & 0.11       & 0.65       \\
			\multicolumn{1}{|l|}{ABIC (bow-free)}
			& {\bf 0.89} & {\bf 0.17} & {\bf 0.72} & {\bf 0.29} & {\bf 0.30} & {\bf 0.45} \\ \hline
		\end{tabular}
		
		\quad
		
		\begin{tabular}{l|ll|ll|ll|}
			\cline{2-7}
			& \multicolumn{2}{c|}{\textsc{Skeleton}}  & \multicolumn{2}{c|}{\textsc{Arrowhead}} & \multicolumn{2}{c|}{\textsc{Tail}}    \\ \hline
			\multicolumn{1}{|l|}{Method}        & tpr $\uparrow$ & fdr $\downarrow$ & tpr $\uparrow$& fdr$\downarrow$ & tpr $\uparrow$& fdr $\downarrow$ \\ \hline
			\multicolumn{1}{|l|}{FCI~\citep{spirtes2000causation}} 
			& 0.51       & 0.12       & 0.41       & 0.53       & 0.10       & 0.73       \\
			\multicolumn{1}{|l|}{gSPo~\citep{bernstein2020ordering}} 
			& {\bf 0.88}       & 0.27       & 0.46       & 0.59       & 0.32       & 0.81       \\
			\multicolumn{1}{|l|}{ABIC (ancestral)}
			& {0.85} & {\bf 0.11} & {\bf 0.72} & {\bf 0.23} & {\bf 0.66} & {\bf 0.47} \\ \hline
		\end{tabular}
	}
	\caption{Comparison of our method to greedyBAP (left) and FCI/greedySPo (right) for recovering 10 variable bow-free and ancestral ADMGs, respectively. We report true positive rate (tpr) and false discovery rate (fdr) --- the fraction of predicted edges that are  present in the target structure or the fraction that are absent from the target structure respectively --- for skeleton, arrowhead and tail recovery. ($\uparrow$/$\downarrow$ indicates higher/lower is better.)}
	\label{tab:results}
\end{table*}

For a given ADMG, we generate data as follows. For each $V_i \rightarrow V_j$ we uniformly sample $\delta_{ij}$ from $\pm[0.5, 2.0],$ for $V_i \leftrightarrow V_j,$ we sample $\beta_{ij}=\beta_{ji}$ from $\pm[0.4, 0.7],$ and for each $\beta_{ii}$ we sample from $\pm[0.7, 1.2]$ and add $\text{sum}(|\beta_{i, -i}|)$ to ensure positive definiteness of $\beta.$

Since randomly generated ADMGs are unlikely to exhibit Verma constraints, we first consider recovery of the ADMG shown in Fig.~\ref{fig:examples}(c) and two other ADMGs $A\rightarrow B\rightarrow C \rightarrow D, B\leftrightarrow D$ and a Markov equivalent ADMG obtained by replacing $A\rightarrow B$ with $A\leftrightarrow B$ which have Verma constraints established in the prior literature. Exact recovery of Fig.~\ref{fig:examples}(c) is possible while the latter ADMGs can be recovered up to ambiguity in the adjacency between $A$ and $B$ as $A\rightarrow B$ or $A\leftrightarrow B.$ We compare our arid and bow-free algorithms to the greedyBAP method proposed in \citep{nowzohour2017distributional} (the only other method available for recovering such constraints). Since greedyBAP is designed to perform random restarts, we allow all methods $5$ uniformly random restarts and pick the final best fitting ADMG. As mentioned earlier, our main hyperparameter is the regularization strength $\lambda,$ which we set to $0.05$ for all experiments.  Choice of other hyperparameters and additional experiments with varying $\lambda$ are provided in Appendix~\ref{app:imp_details},~\ref{app:exprs}. We generate $100$ datasets for each sample size of $[500, 1000, 1500, 2000]$ from a uniform sample of the $3$ aforementioned ADMGs. The results are summarized via barplots in Fig.~\ref{fig:experiments}.

The ABIC arid and bow-free procedures both outperform the greedyBAP procedure in recovering the true equivalence class. The highest recovery rate is shown by the bow-free procedure with $39\%$ at $n=1000.$ Though this seems low, these results are quite promising in light of geometric arguments in \cite{evans2018model} that show reliable recovery of Verma constraints may require very large sample sizes.  In examining the modes of failure of each algorithm, our ABIC procedures often fail to recover the true ADMG by returning a super model of the true equivalence class while the greedyBAP procedure often returns an incorrect independence model; see Fig.~\ref{fig:experiments_super} in Appendix~\ref{app:exprs}. The former kind of mistake does not yield bias in downstream inference tasks while the latter does. Our bow-free procedure yields more accurate results than the arid one most likely due to posing an easier optimization problem. In the $400$ runs used to generate plots in Fig.~\ref{fig:experiments}, the bow-free procedure failed to converge only $3$ times and the arid one never failed to converge, which is consistent with established theoretical results on almost-everywhere and global identifiability of these models.

For larger randomly generated arid ADMGs, to save computation time, we only compare our bow-free procedure with greedyBAP, and for ancestral ADMGs, we compare our ancestral procedure with FCI \citep{spirtes2000causation} and greedySPo \citep{bernstein2020ordering}. We also obtained results for GFCI \citep{ogarrio2016hybrid} and M3HC \citep{tsirlis2018scoring}. These were slightly worse than the results for FCI and greedySPo so we only report the latter results. Runs of the M3HC algorithm typically ended with convergence warnings.\footnote{Code from https://github.com/mensxmachina/M3HC.} Random arid/ancestral ADMGs on $10$ and $15$ variables were generated by first producing a random bow-free ADMG with directed and bidirected edge probabilities of $0.4$ and $0.3$ respectively, and then applying the maximal arid/ancestral projection. We report true positive and false discovery rates for exact skeleton recovery of the true ADMG as well as recovery of tails and arrowheads in the true PAG for $100$ datasets of $1000$ samples each. For FCI, we used a significance level of $0.15$ which gave the most competitive results.  Our method performs favorably in recovery of both arid and ancestral ADMGs. 
Results for $10$ variables, which roughly matches the dimensionality of our data application, are summarized in Table~\ref{tab:results}. Results for $15$ variables showing the same trends are in Appendix~\ref{app:exprs}.

Finally we apply our ABIC bow-free method to a cleaned version of the protein expression dataset in \cite{sachs2005causal} from \cite{ramsey2018fask}. The result is shown in the right panel of Fig.~\ref{fig:experiments}. The precision and recall of our procedure with respect to the true adjacencies provided in \cite{ramsey2018fask} are $0.77$ and $0.61$ respectively. We do not provide evaluation of orientations as there is no consensus regarding many of them. However, we briefly highlight the importance of a Verma restriction in producing a model that is consistent with an intervention experiment performed by \cite{sachs2005causal}.
The authors found that manipulation of Erk produced no downstream effect on PKA though they are correlated.
The ADMG in Fig.~\ref{fig:experiments} has an edge $\text{Erk} \leftrightarrow \text{PKA}$ that is consistent with this finding. Moreover, this edge cannot be oriented in either direction without producing different independence models than the one implied by Fig.~\ref{fig:experiments}. This is due to a Verma restriction between Akt and PKC; we provide more details in Appendix~\ref{app:protein}. We confirm that orienting the edge as $\text{Erk} \leftarrow \text{PKA}$ or $\text{Erk} \rightarrow \text{PKA}$ leads to an increase in the BIC score, indicating that the Verma restriction capturing the ground truth is preferred over these other explanations.

\section{CONCLUSION}
\label{sec:conclusion}

We have extended the continuous optimization scheme of causal discovery to include models that capture all equality constraints on the observed margin of hidden variable linear SEMs with Gaussian errors. The differentiable algebraic constraints we provided are non-parametric and may thus enable future development of non-parametric causal discovery methods. Our method may also help explore questions regarding distributional equivalence and Markov equivalence with respect to all equality restrictions in ADMG models. The authors in \cite{shpitser2014introduction} made progress on equivalence theory for 4-variable ADMGs by enumerating all possible 4-variable ADMGs and evaluating the BIC score for each one, grouping graphs with equal scores to form an ``empirical equivalence class.'' A similar approach could be pursued for larger graphs using our proposed causal discovery procedure. If relevant patterns in larger empirical equivalence classes become apparent, this may result in progress towards a characterization for nested Markov equivalence.

\subsubsection*{Acknowledgements}
\label{sec:acks}

The authors would like to thank Razieh Nabi for her insightful comments regarding approximations of the Bayesian Information Criterion. This project is sponsored in part by the NSF CAREER grant 1942239. The content of the information does not necessarily reflect the
position or the policy of the Government, and no official
endorsement should be inferred.

\clearpage

\onecolumn
\appendix
\setcounter{figure}{0}
\setcounter{table}{0}

\begin{center}
	\LARGE{\textbf{Appendix:  Differentiable Causal Discovery Under Unmeasured Confounding }}
\end{center}
\vspace{0.75cm}
\renewcommand{\thetable}{\Alph{table}}
\renewcommand{\thefigure}{\Alph{figure}}

The Appendix is organized as follows. In Appendix~\ref{app:greenery} we discuss details of the \textproc{Greenery} algorithm for penalizing c-trees and introduce the formalizations necessary to prove its correctness. In Appendix~\ref{app:protein} we provide additional comments on the protein expression network learned by applying our method to the data from \cite{sachs2005causal}. In Appendix~\ref{app:proofs} we present formal proofs of results in our paper. In Appendix~\ref{app:imp_details} we discuss additional implementation details and choice of hyperparameters for our experiments. Finally in Appendix~\ref{app:exprs} we provide additional experiments not included in the main draft of the paper.

\section{DETAILS OF THE GREENERY ALGORITHM}
\label{app:greenery}

\cite{bhattacharya2020semiparametric} introduced a graphical and probabilistic operator called \emph{primal fixing} that can be applied recursively to an ADMG and its statistical model to identify causal parameters of interest. In this section we provide the necessary background on the graphical operator and discuss how it relates to the detection of c-trees. We then show how primal fixing is codified in the steps of Algorithm~\ref{alg:greenery} through an example.

A conditional ADMG (CADMG) $\G=(V, W, E)$ is an ADMG whose vertices can be partitioned into random vertices $V$ and fixed vertices $W,$ with the restriction that no arrowheads point into $W$ \citep{richardson2017nested}. A vertex $V_i$ in a CADMG $\G=(V, W, E)$ is said to be primal fixable if there is no bidirected path from $V_i$ to any of its direct children. The graphical operation of primal fixing  $V_i$ in $\G,$ denoted by $\phi_{V_i}(\G),$ yields a new CADMG $\G=(V\setminus V_i, W \cup V_i, E\setminus \{e \in E \mid e = \circ \rightarrow V_i \text{ or } \circ \leftrightarrow V_i\})$ where $V_i$ is now ``fixed'' (denoted by a square box in figures shown in this Supplement) and incoming edges into $V_i$ are deleted. This can be extended to a set of vertices as follows. A set of $k$ vertices $S$ is said to be primal fixable if there exists an ordering $(S_1, \dots, S_{k})$  such that $S_1$ is primal fixable in $\G,$ $S_2$ is primal fixable in $\phi_{S_1}(\G),$ $S_3$ is primal fixable in $\phi_{S_2}(\phi_{S_1}(\G)),$ and so on. It is easy to see that any such valid ordering on $S$ yields the same final CADMG. Hence, we can denote primal fixing a set of vertices $S$ as simply $\phi_S(\G).$ A vertex $V_i$ in an ADMG $\G$ is said to be \emph{reachable} if $V\setminus V_i$ is primal fixable in $\G.$ \cite{shpitser2018acyclic} showed that if $V_i$ is reachable in $\G,$ then the causal effect of the parents of $V_i$ on $V_i$ itself is identified, and there is no $V_i$ rooted c-tree in $\G.$\footnote{Actually this was shown with respect to the ordinary fixing operator proposed in \cite{richardson2017nested} which performs the same graphical operation as primal fixing but considers $V_i$ to be fixable when there are no bidirected paths to any descendant (a vertex $V_j$ such that there exists a directed path from $V_i$ to $V_j$) of $V_i.$ It is easy to see how primal fixing is a strict generalization of fixing by noting that the children of $V_i$ is a subset of its descendants.} If no valid primal fixing order exists, $V_i$ along with the unique minimal set of vertices that could not be primal fixed form a $V_i$-rooted c-tree \citep{shpitser2018acyclic}. That is, an ADMG $\G$ is arid if and only if every vertex $V_i \in V$ is reachable. This forms the basis of Algorithm~\ref{alg:greenery}.

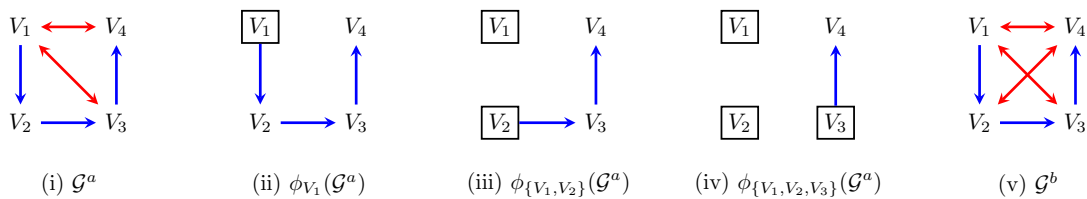
\begin{figure}[h]
	\begin{center}
		\scalebox{0.85}{
			\begin{tikzpicture}[>=stealth, node distance=1.5cm]
			\tikzstyle{format} = [thick, circle, minimum size=1.0mm, inner sep=0pt]
			\tikzstyle{square} = [draw, thick, minimum size=1.0mm, inner sep=3pt]
			
			\begin{scope}[xshift=0cm]
			\path[->, very thick]
			node[] (a) {$V_1$}
			node[right of=a] (b) {$V_4$}
			node[below of=a] (c) {$V_2$}
			node[below of=b] (d) {$V_3$}
			node[below right of=c, xshift=-0.3cm, yshift=0.1cm] (label) {(i) $\G^a$}
			
			(a) edge[blue] (c)
			(c) edge[blue] (d)
			(d) edge[blue] (b)
			(a) edge[red, <->] (b)
			(a) edge[red, <->] (d)
			;
			\end{scope}
			
			\begin{scope}[xshift=3.75cm]
			\path[->, very thick]
			node[square] (a) {$V_1$}
			node[right of=a] (b) {$V_4$}
			node[below of=a] (c) {$V_2$}
			node[below of=b] (d) {$V_3$}
			node[below right of=c, xshift=-0.3cm, yshift=0.1cm] (label) {(ii) $\phi_{V_1}(\G^a)$}
			
			(a) edge[blue] (c)
			(c) edge[blue] (d)
			(d) edge[blue] (b)
			;
			\end{scope}
			
			\begin{scope}[xshift=7.5cm]
			\path[->, very thick]
			node[square] (a) {$V_1$}
			node[right of=a] (b) {$V_4$}
			node[square, below of=a] (c) {$V_2$}
			node[below of=b] (d) {$V_3$}
			node[below right of=c, xshift=-0.3cm, yshift=0.1cm] (label) {(iii) $\phi_{\{V_1,V_2\}}(\G^a)$}
			
			(d) edge[blue] (b)
			(c) edge[blue] (d)
			;
			\end{scope}
			
			\begin{scope}[xshift=11.25cm]
			\path[->, very thick]
			node[square] (a) {$V_1$}
			node[right of=a] (b) {$V_4$}
			node[square, below of=a] (c) {$V_2$}
			node[square, below of=b] (d) {$V_3$}
			node[below right of=c, xshift=-0.3cm, yshift=0.1cm] (label) {(iv) $\phi_{\{V_1,V_2,V_3\}}(\G^a)$}
			
			(d) edge[blue] (b)
			;
			\end{scope}

			\begin{scope}[xshift=15cm]
			\path[->, very thick]
			node[] (a) {$V_1$}
			node[right of=a] (b) {$V_4$}
			node[below of=a] (c) {$V_2$}
			node[below of=b] (d) {$V_3$}
			node[below right of=c, xshift=-0.3cm, yshift=0.1cm] (label) {(v) $\G^{b}$}
			
			(a) edge[blue] (c)
			(c) edge[blue] (d)
			(d) edge[blue] (b)
			(a) edge[red, <->] (b)
			(a) edge[red, <->] (d)
			(c) edge[red, <->] (b)
			;
			\end{scope}
			
			\end{tikzpicture}
		}
	\end{center}
	\caption{(i) An arid ADMG; (ii) The CADMG obtained after primal fixing $V_1;$ (iii) The CADMG obtained after primal fixing $V_1$ and $V_2;$ (iv) The CADMG obtained after primal fixing $V_1, V_2, $ and $V_3;$ (v) A non-arid bow-free ADMG that is a super model of (i).}
	\label{fig:supp_examples}
\end{figure}


We now demonstrate usage of the primal fixing operator to establish that the ADMG $\G^a$ shown in Fig.~\ref{fig:supp_examples}(i) is arid and the ADMG $\G^b$ shown in Fig.~\ref{fig:supp_examples}(v) is not. These are the same graphs shown in Section~\ref{sec:motiv} of the paper but we redraw and relabel them here for convenience. The reachability of vertices $V_1, V_2,$ and $V_3$ in $\G^a$ is easily established. In every case, we can primal fix the remaining vertices in a reverse topological order starting with $V_4$ which has no children. The reachability of $V_4$ is established by noticing that  $V_1$ is primal fixable in $\G^a.$ In the resulting CADMG, shown in Fig.~\ref{fig:supp_examples}(ii), both $V_2$ and $V_3$ are primal fixable. Primal fixing $V_2$ yields the CADMG in Fig.~\ref{fig:supp_examples}(iii) and finally primal fixing $V_3$ yields the CADMG in Fig.~\ref{fig:supp_examples}(iv). Hence, all vertices in $\G^a$ are reachable. It then follows that $\G^a$ is arid. If we try to apply the same reasoning to the $\G^b$ in Fig.~\ref{fig:supp_examples}(v), we see that $V_1, V_2,$ and $V_3$ are still reachable as before. However, we cannot establish a sequence of primal fixing operations to reach $V_4$ as none of the other vertices are primal fixable in the original graph. Hence, there is a $V_4$-rooted c-tree in $\G^b$ comprised of the arborescence $V_1\rightarrow V_2 \rightarrow V_3 \rightarrow V_4$ which also forms a bidirected component in $\G^b.$

\subsection{Example Application of the Greenery Algorithm}

We now demonstrate how the above primal fixing steps relate to Algorithm~\ref{alg:greenery}. Let the ordering of vertices of entries in the matrix be $V_1, V_2, V_3, V_4.$ The adjacency matrices $D$ and $B$ for $\G^a$ in Fig.~\ref{fig:supp_examples}(i) are as follows.
\begin{align*}
	D=\begin{bmatrix}
		0  &  1 & 0 & 0 \\
		0  &  0 & 1 & 0	\\
		0  &  0 & 0 & 1	\\
		0  &  0 & 0 & 0
	\end{bmatrix}
	\qquad  \qquad
	B=\begin{bmatrix}
		0  &  0 & 1 & 1 \\
		0  &  0 & 0 & 0	\\
		1  &  0 & 0 & 0	\\
		1  &  0 & 0 & 0
	\end{bmatrix}.
\end{align*}
The $i^{th}$ iteration of the outer loop of the algorithm attempts to establish the reachability of $V_i$, and hence, the presence or absence of a $V_i$-rooted c-tree. Note that since the primal fixing operation can be applied at most $d-1$ times (where $d$ is the number of vertices in $\G$) to determine the reachability of $V_i,$ the inner loop of Algorithm~\ref{alg:greenery} also executes $d-1$ times.
We now focus on the final iteration of the algorithm where it tries to establish the reachability of $V_4.$

In the first iteration of the inner loop we have $D^f=D$ and $B^f=B.$ Therefore we have,
\begin{align*}
	e^{B^f} \circ D=\begin{bmatrix}
		0  &  0 & 0 & 0 \\
		0  &  0 & 0 & 0	\\
		0  &  0 & 0.59 & 0	\\
		0  &  0 & 0 & 0
	\end{bmatrix}
	\qquad  \qquad
	f=\begin{bmatrix}
		0 & 0 & 0.53 & 0.76 
	\end{bmatrix}
	\qquad  \qquad
	F=\begin{bmatrix}
		0  &  0 & 0.53 & 0.76 \\
		0  &  0 & 0.53 & 0.76	\\
		0  &  0 & 0.53 & 0.76	\\
		0  &  0 & 0.53 & 0.76
	\end{bmatrix}.
\end{align*}
Each entry $i, j$ of the matrix $e^{B_f}\circ D$ is zero if and only if a bidirected path from $V_i$ to $V_j$ and a directed edge $V_i \rightarrow V_j$ do not co-exist in $\G.$ The sum of the $i^{th}$ row of this matrix then exactly characterizes the primal fixability criterion. That is, $V_i$ is primal fixable if and only if the sum of the $i^{th}$ row in $e^{B_f}\circ D$ is $0.$ The above calculations indicate that the vertices $V_1, V_2,$ and $V_4$ are all primal fixable in $\G^a,$ which can be easily confirmed by looking at the graph itself. The vector $f$ then summarizes the primal fixability of each vertex except we add the $i^{th}$ row of an identity matrix to ensure that we do not accidentally primal fix $V_i$ itself when determining its reachability. The matrix $F$ formed by tiling the $f$ vector $d$ times can then be used as a ``mask'' that implements the primal fixing operation applied to $V_1$ and $V_2$ simultaneously, yielding the following updates to $D^f$ and $B^f.$
\begin{align*}
	D^f=\begin{bmatrix}
		0  &  0 & 0 & 0 \\
		0  &  0 & 0.53 & 0	\\
		0  &  0 & 0 & 0.76	\\
		0  &  0 & 0 & 0
	\end{bmatrix}
	\qquad  \qquad
	B^f=\begin{bmatrix}
		0  &  0 & 0 & 0 \\
		0  &  0 & 0 & 0	\\
		0  &  0 & 0 & 0	\\
		0  &  0 & 0 & 0
	\end{bmatrix}.
\end{align*}
It is easy to confirm that the induced ADMG $\G(D^f, B^f)$ corresponds to the CADMG shown in Fig.~\ref{fig:supp_examples}(iii). Note that a constant positive scaling factor can also be applied to the hyperbolic tangent function to improve the sharpness of the approximation of the primal fixing operator. In the second iteration of the loop, we apply the same process again and obtain,
\begin{align*}
	e^{B^f} \circ D=\begin{bmatrix}
		0  &  0 & 0 & 0 \\
		0  &  0 & 0 & 0	\\
		0  &  0 & 0 & 0	\\
		0  &  0 & 0 & 0
	\end{bmatrix}
	\qquad  \qquad
	f=\begin{bmatrix}
		0 & 0 & 0 & 0.76 
	\end{bmatrix}
	\qquad  \qquad
	F=\begin{bmatrix}
		0  &  0 & 0 & 0.76 \\
		0  &  0 & 0 & 0.76	\\
		0  &  0 & 0 & 0.76	\\
		0  &  0 & 0 & 0.76
	\end{bmatrix}.
\end{align*}
That is, in the second iteration of the algorithm, $V_3$ becomes primal fixable. Applying the primal fixing operator yields the adjacency matrices,
\begin{align*}
	D^f=\begin{bmatrix}
		0  &  0 & 0 & 0 \\
		0  &  0 & 0 & 0	\\
		0  &  0 & 0 & 0.58	\\
		0  &  0 & 0 & 0
	\end{bmatrix}
	\qquad  \qquad
	B^f=\begin{bmatrix}
		0  &  0 & 0 & 0 \\
		0  &  0 & 0 & 0	\\
		0  &  0 & 0 & 0	\\
		0  &  0 & 0 & 0
	\end{bmatrix},
\end{align*}
which induce the CADMG shown in Fig.~\ref{fig:supp_examples}(iv) corresponding to primal fixing $V_3.$ Thus, in this case, reachability of $V_4$ is established in $2$ steps. However, the algorithm will still perform a third step that does not result in any additional primal fixing and does not change the conclusion of  reachability of $V_4.$ As there are no vertices that have both a bidirected path and directed path to $V_4$ in the final CADMG and corresponding adjacency matrices, $C=e^{B^f}\circ e^{D^f}$ is simply the identity matrix. Taking the $i^{th}$ column sum then evaluates to $1$ which is subtracted off later in the final ``return'' step of the algorithm. A similar argument holds for vertices $V_1, V_2,$ and $V_3.$ Thus, applying Algorithm~\ref{alg:greenery} to $\G^a$ in Fig.~\ref{fig:supp_examples}(i) returns a value of $0$ confirming that $\G^a$ is arid. 

We now consider application of the algorithm to the ADMG $\G^b$ shown in Fig.~\ref{fig:supp_examples}(v). We will apply a scaling constant of $10$ to the hyperbolic tangent function, i.e., we use $\text{tanh}(10x),$ so that the values are large enough to illustrate the main concept. We again focus on the reachability of $V_4.$ The adjacency matrices for $\G^b$ are:
\begin{align*}
	D=\begin{bmatrix}
		0  &  1 & 0 & 0 \\
		0  &  0 & 1 & 0	\\
		0  &  0 & 0 & 1	\\
		0  &  0 & 0 & 0
	\end{bmatrix}
	\qquad  \qquad
	B=\begin{bmatrix}
		0  &  0 & 1 & 1 \\
		0  &  0 & 0 & 1	\\
		1  &  0 & 0 & 0	\\
		1  &  1 & 0 & 0
	\end{bmatrix}.
\end{align*}
In the first iteration of the inner loop we have,
\begin{align*}
	e^{B^f} \circ D=\begin{bmatrix}
		0  &  0.64 & 0 & 0 \\
		0  &  0 & 0.19 & 0	\\
		0  &  0 & 0 & 0.64	\\
		0  &  0 & 0 & 0
	\end{bmatrix}
	\qquad  \qquad
	f=\begin{bmatrix}
		1 & 0.96 & 1 & 1 
	\end{bmatrix}
	\qquad  \qquad
	F=\begin{bmatrix}
		1  &  0.96 & 1 & 1 \\
		1  &  0.96 & 1 & 1	\\
		1  &  0.96 & 1 & 1	\\
		1  &  0.96 & 1 & 1
	\end{bmatrix}.
\end{align*}
That is, we see that none of the vertices in $\G^b$ are primal fixable. Therefore applying the primal fixable operator through the matrix $F$ results in adjacency matrices,
\begin{align*}
	D^f=\begin{bmatrix}
		0  &  0.96 & 0 & 0 \\
		0  &  0 & 1 & 0	\\
		0  &  0 & 0 & 1	\\
		0  &  0 & 0 & 0
	\end{bmatrix}
	\qquad  \qquad
	B^f=\begin{bmatrix}
		0  &  0 & 1 & 1 \\
		0  &  0 & 0 & 0.96	\\
		1  &  0 & 0 & 0	\\
		1  &  0.96 & 0 & 0
	\end{bmatrix},
\end{align*}
which induce a ``CADMG'' that has the same edges as the original graph $\G^b.$ Repeated applications of this in the second and third iterations do not change the structure of the induced graph. Therefore, upon termination of the inner loop, there remains a directed path from every vertex in $V\setminus V_4$ to $V_4$ and the vertices still form a bidirected connected component. That is, there is a $V_4$-rooted c-tree in $\G^b.$ This is confirmed when we evaluate the sum of the $i^{th}$ column of $C=e^{D^f}\circ e^{B^f}$ to $2.34.$ The other vertices $V_1, V_2,$ and $V_3$ are still reachable and their respective column sums upon termination of the inner loop yield a value of $1$ each. Subtracting $d$ at the end of the algorithm still leaves a positive remainder of $1.34.$ Hence, Algorithm~\ref{alg:greenery} returns a  positive quantity when applied to $\G^b,$ confirming that it is not arid.

\clearpage

\section{COMMENTS ON PROTEIN EXPRESSION ANALYSIS}
\label{app:protein}

In this section we discuss the Verma restriction that allows us to establish that Erk is \emph{not} a cause of PKA. The importance of this relation stems from manipulation of Erk by the authors of \cite{sachs2005causal} and establishing that no downstream change was observed in PKA.

We first point out that there is no ordinary conditional independence constraint between Akt and PKC in the learned structure shown in the right panel of Fig.~\ref{fig:experiments}, despite the absence of an edge between the two. This can be confirmed by noting the presence of an \emph{inducing path} between Akt and PKC. An inducing path between $V_i$ and $V_j$ is a path from $V_i$ to $V_j$ where every non-endpoint is both a collider ($\rightarrow \circ \leftarrow, \leftrightarrow \circ \leftarrow,$ or $\leftrightarrow \circ \leftrightarrow$) and has a directed path to either $V_i$ or $V_j.$ It is well-known that the presence of such a path precludes the possibility of an ordinary conditional independence of the form $V_i \ci V_j \mid Z$ for any $Z \subseteq V\setminus \{V_i, V_j\}$ \citep{verma1990equivalence}. In our analysis it can be confirmed that $\text{Akt}\rightarrow \text{Erk}\leftrightarrow \text{PKA} \leftrightarrow \text{PKC}$ is an inducing path between Akt and PKC. Thus, there is no ordinary conditional independence between these two proteins under our learned model. However, under the faithfulness assumption, the absence of the edge between Akt and PKC implies an equality restriction. We now provide the non-parametric form of the corresponding Verma constraint.

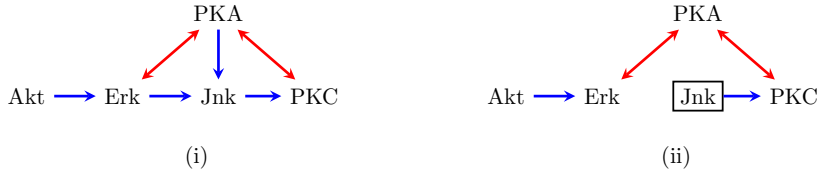
\begin{figure}[t]
	\begin{center}
		\scalebox{0.85}{
			\begin{tikzpicture}[>=stealth, node distance=1.5cm]
			\tikzstyle{format} = [thick, circle, minimum size=1.0mm, inner sep=0pt]
			\tikzstyle{square} = [draw, thick, minimum size=1.0mm, inner sep=3pt]
			
			\begin{scope}[xshift=0cm]
			\path[->, very thick]
			node[] (akt) {Akt}
			node[right of=akt] (erk) {Erk}
			node[right of=erk] (jnk) {Jnk}
			node[right of=jnk] (pkc) {PKC}
			node[above of=jnk, yshift=-0.2cm] (pka) {PKA}
			node[below right of=erk, xshift=0.1cm, yshift=0.1cm] (label) {(i)}
			
			(akt) edge[blue] (erk)
			(erk) edge[blue] (jnk)
			(jnk) edge[blue] (pkc)
			(pka) edge[blue] (jnk)
			(erk) edge[red, <->] (pka)
			(pka) edge[red, <->] (pkc)
			;
			\end{scope}
			
			\begin{scope}[xshift=7.5cm]
			\path[->, very thick]
			node[] (akt) {Akt}
			node[right of=akt] (erk) {Erk}
			node[square, right of=erk] (jnk) {Jnk}
			node[right of=jnk] (pkc) {PKC}
			node[above of=jnk, yshift=-0.2cm] (pka) {PKA}
			node[below right of=erk, xshift=0.1cm, yshift=0.1cm] (label) {(ii)}
			
			(akt) edge[blue] (erk)
			(jnk) edge[blue] (pkc)
			(erk) edge[red, <->] (pka)
			(pka) edge[red, <->] (pkc)
			;
			\end{scope}
			
			\end{tikzpicture}
		}
	\end{center}
	\caption{(i) A subgraph of the protein network in Fig.~\ref{fig:experiments} that we use to highlight the Verma constraint between Akt and PKC; (ii) A CADMG corresponding to the post-intervention distribution that would be obtained by intervening on Jnk. }
	\label{fig:supp_protein}
\end{figure}

Consider the ADMG and corresponding distribution obtained by recursively marginalizing out all vertices (except PKC) with no outgoing directed edges in Fig.~\ref{fig:experiments}. In performing this graphical operation, none of the variables removed act as a latent confounder for the remaining variables in the problem. Therefore, by rules of latent projection described in \cite{verma1990equivalence}, we simply obtain a subgraph of the original network as shown in Fig.~\ref{fig:supp_protein}(i). Note that the inducing path between Akt and PKC is still preserved. Let $p(V^s)$ be the corresponding marginal distribution on the remaining subset of variables. The Verma constraint is then given by,
\begin{align*}
	\text{Akt} \ci \text{PKC} \text{ in }\  \frac{p(V^s)}{p(\text{Jnk} \mid \text{Erk}, \text{PKA})}.
\end{align*}
Intuitively, one can view the independence between Akt and PKC as manifesting in a post-intervention distribution obtained after intervening on Jnk, resulting in the CADMG (or truncated ADMG) shown in Fig.~\ref{fig:supp_protein}(ii) where incoming edges to Jnk are removed. The resulting independence is then easily read off from the CADMG via the m-separation criterion \citep{richardson2003markov}. See \cite{tian2002testable} and \cite{richardson2017nested} for more details on how to derive such constraints in general. Orienting the $\text{Erk} \leftrightarrow \text{PKA}$ edge as either $\text{Erk} \leftarrow \text{PKA}$ or $\text{Erk} \rightarrow \text{PKA}$ breaks the inducing path between Akt and PKC, meaning that either orientation produces a different independence model implying an ordinary independence constraint instead of the Verma restriction. We evaluated the BIC scores with either orientation and confirm that they both yield an increase in the score. This indicates that our learned model which posits that Erk is correlated with PKA through unmeasured confounding is the preferred causal explanation. This explanation is consistent with experiments performed in \cite{sachs2005causal}, and we are able to arrive at the same conclusion from purely observational data. Moreover, this explanation was differentiated from others via the Verma restriction between Akt and PKC, highlighting the value of considering general equality restrictions beyond ordinary conditional independence.

\clearpage

\section{PROOFS}
\label{app:proofs}

\begin{thma}{\ref{thm:constraints}}
	The constraints shown in Table~\ref{tab:constraints} are satisfied if and only if the adjacency matrices satisfy the relevant property of ancestrality, aridity, and bow-freeness respectively.
\end{thma}

\begin{proof}
	We use the following facts for all of our proofs.
	The matrix exponential of a square matrix $A$ is defined as the infinite Taylor series, 
	\begin{align}
		e^A = \sum_{k=0}^{\infty} \frac{1}{k!} A^k.
	\end{align}
	For a binary square matrix $A,$ corresponding to a directed/bidirected adjacency matrix, the entry $A_{ij}^k$ counts the number of directed/bidirected walks of length $k$ from vertex $i$ to vertex $j;$ see for example \citep{butler2008eigenvalues}.
	
	\subsubsection*{Ancestral ADMGs}
	Consider the constraint shown in Table~\ref{tab:constraints}. That is,
	\begin{align*}
		\trace{e^D} - d + \sumall{e^D \circ B} = 0.
	\end{align*}
	It is easy to see from results in \citep{zheng2018dags} that the constraint $\trace{e^D} - d=0$ is satisfied if and only if the induced graph $\G(D, B)$ is acyclic.
	We now show that $\sumall{e^D \circ B}=0$ if and only if $\G$ is ancestral.
	
	By definition of the matrix exponential,
	\begin{align*}
		\sumall{e^D \circ B} &= \text{sum}\bigg(I\circ B + \sum_{k=1}^{\infty} \frac{1}{k!}D^k \circ B\bigg) \\
		&= \text{sum}\bigg(I\circ B\bigg) + \sum_{k=1}^{\infty} \frac{1}{k!}\text{sum}\bigg( D^k \circ B\bigg),
	\end{align*}
	where the second equality follows from basic matrix properties.
	
	The first term in the series, $\sumall{I \circ B},$ counts the number of self bidirected edges $V_i \leftrightarrow V_i$ which is a special-case violation of ancestrality. This term is zero if no such edges exist. An entry $i, j$ in the matrix $D^k \circ B$ counts the number of occurences of directed paths from $V_i$ to $V_j$ of length $k$ such that $V_i$ and $V_j$ are also connected via a bidirected edge. Therefore, all remaining terms of the form $\frac{1}{k!}\sumall{D^k \circ B}$ count the number of directed paths of length $k$ that violate the ancestrality property rescaled by a positive factor of $\frac{1}{k!}.$ That is, these terms are all $\geq 0$ and equal to zero only when no such paths exist, i.e., $\G$ is ancestral.
	
	\subsubsection*{Arid ADMGs}
	Consider the constraint shown in Table~\ref{tab:constraints}. That is,
	\begin{align*}
		\trace{e^D} - d + \textproc{Greenery}(D,B) = 0.
	\end{align*}
	The terms $	\trace{e^D} - d$ capture the acyclicity constraint as before. We now show that the output of Algorithm~\ref{alg:greenery} is zero if and only if $\G$ satisfies the arid property. That is, $\textproc{Greenery}(D, B)=0$ is satisfied if and only if $\G$ is arid. The background required for this proof was laid out in Appendix~\ref{app:greenery}.
	
	The outer loop of Algorithm~\ref{alg:greenery} iterates over each vertex $V_i$ in order to evaluate its reachability, or equivalently, the presence/absence of a $V_i$-rooted c-tree \citep{shpitser2018acyclic}. The inner loop achieves this as follows.
	
	Reachability of $V_i$ can be determined in at most $d-1$ primal fixing operations. Therefore, the inner loop executes $d-1$ times. On each iteration, the algorithm considers the primal fixability of vertices by effectively treating the matrices $D^f$ and $B^f$ as adjacency matrices of a CADMG. In the first iteration, $D^f$ and $B^f$ are initialized with values from the directed and bidirected adjacency matrices respectively. The sum of the $j^{th}$ row in the matrix $e^{B^f}\circ D^f$ evaluates to zero if and only if there are no bidirected paths from $V_j$ to any of its direct children $V_k,$ which exactly corresponds to the graphical criterion for determining primal fixability of $V_j.$ The addition of the $i^{th}$ row of an identity matrix to $t$ ensures that $V_i$ itself is not treated as primal fixable when evaluating its reachability. Therefore, in the first iteration, the vector $f$ encodes a smoothened version (due to the application of the hyperbolic tangent function) of the usual primal fixability criterion for all vertices $V\setminus V_i$ in the original graph $\G.$ Tiling the vector $f$ to form the $d\times d$ matrix $F$ allows us to apply the softened version of primal fixing to the adjacency matrices, which is performed in lines~\ref{alg:greenery_fix}-\ref{alg:greenery_fix_end} of the algorithm. On the next iteration, the matrices $D^f$ and $B^f$ can then be treated as adjacency matrices of a CADMG obtained by primal fixing a set of vertices, say $S_1,$ that satisfied the primal fixability criterion in $\G.$ The same logic can be applied to subsequent iterations of the algorithm where we determine the primal fixability of a set of vertices $V\setminus (S_1\cup V_i)$ in $\phi_{S_1}(\G),$ denote the primal fixable vertices as $S_2,$ and then proceed to do the same for $V\setminus (S_1 \cup S_2 \cup V_i)$ in $\phi_{S_1 \cup S_2}(\G),$ and so on.
	
	On termination of the inner loop, we have that $S_1 \cup S_2, \dots, \cup S_{d-1} \subseteq V\setminus V_i.$ We first consider the case when equality holds. In this case, $V_i$ is reachable, from which it follows that there is no $V_i$-rooted c-tree in $\G$ \citep{shpitser2018acyclic}. The final matrices $D^f$ and $B^f$ then correspond to a CADMG where all vertices except $V_i$ have been primal fixed. In such a CADMG the only edges that may be present are directed edges into $V_i$ due to the removal of incoming edges to all other vertices in the graph. Thus, $e^{B^f}$ evaluates to an identity matrix as there are no bidirected edges. Assuming $\G$ is a graph with no directed cycles (which is already enforced by the first two terms in the arid constraint), the Hadamard product $C=e^{B^f}\circ e^{D^f}$ is then also an identity matrix. Taking the sum of the $i^{th}$ column of $C$ then simply evaluates to $1.$ If every vertex $V_i \in V$ is reachable in this manner, it implies that the graph is arid, and the greenery quantity will then evaluate to $d.$ The subtraction of $d$ in the ``return'' statement of Algorithm~\ref{alg:greenery} then returns a value of $0$ for arid graphs. Now we consider the case when equality does not hold, i.e., there exists a set of vertices $X = V\setminus V_i \setminus (S_1 \cup S_2 \dots \cup S_{d-1})$ that could not be primal fixed. This implies that $V_i$ is not reachable and there exists a $V_i$-rooted c-tree. By definition, the structure of this c-tree comprises of directed and bidirected paths from vertices in $X$ to $V_i.$ The sum of the $i^{th}$ column in $C=e^{B^f}\circ e^{D^f}$ then provides a weighted count of these paths. Subtracting off $d$ in the final ``return'' statement then yields a  positive quantity that provides a weight for each $V_i$-rooted c-tree detected in a non-arid graph $\G.$

	\subsubsection*{Bow-free ADMGs}
	Consider the constraint shown in Table~\ref{tab:constraints}. That is,
	\begin{align*}
		\trace{e^D} - d + \sumall{D \circ B} = 0.
	\end{align*}
	The terms $	\trace{e^D} - d$ capture the acyclicity constraint as before. It is easy to see that the term $\sumall{D \circ B}$ counts the number of bows in the induced graph $\G.$ Hence, $\sumall{D \circ B}$ is zero if and only if $\G$ is bow-free.
	
\end{proof}

\begin{thma}{\ref{lem:optimum}}
	Let $p(V; \theta^*)$ be a distribution in the curved exponential family that is Markov and faithful  with respect to an arid ADMG $\G^*$. Finding the global optimum of the continuous program in display~(\ref{eq:reformulation}) with $f\equiv BIC$ yields an ADMG $\G(\theta)$ that implies the same equality restrictions as $\G^*.$
\end{thma}

\begin{proof}
	This follows immediately from the validity of the constraints in Theorem~\ref{thm:constraints} and the consistency of the BIC score for model selection in curved exponential families \citep{haughton1988on}.
	
\end{proof}

\begin{cora}{\ref{cor:constraints_power}}
	The results in Theorem~\ref{thm:constraints} and Corollary~\ref{cor:constraints_sem} hold if every occurrence of a matrix exponential $e^{A}$ is replaced with the matrix power $(I + c A)^d$ for any $c > 0,$ where $I$ is the identity matrix.
\end{cora}

\begin{proof}
	The proof is straightforward by noting that the binomial expansion of $(I + c A)^d=I + \sum_{k=1}^d \binom{d}{k} c^k A^k$ which is similar to the infinite series expansion of the matrix exponential truncated to $d$ terms. As paths greater than length $d$ are irrelevant in a system with $d$ vertices, these terms are sufficient.
	
\end{proof}

\clearpage

\section{IMPLEMENTATION DETAILS}
\label{app:imp_details}

In this section we discuss implementation details of our procedure that were not included in the main paper.

\subsection*{Implementation of Constraints}

As mentioned in the main paper, we use the representation of constraints in Table~\ref{tab:constraints} obtained by replacing each matrix exponential $e^A$ with $(I + cA)^d.$ We have two primary reasons for doing so. First, as pointed out by \cite{yu2019dag}, the latter representation is numerically more stable. Second, by evaluating the binomial expansion $(I + cA)^d=I + \sum_{k=1}^d \binom{d}{k} c^k A^k$ explicitly, we are able to obtain analytic gradients for our constraints automatically via the HIPS Autograd package \citep{maclaurin2015autograd, maclaurin2016modeling}. Analytic gradients for the matrix exponential on the other hand are not easily obtained and the function itself is not implemented in many popular computing libraries. In our implementation we use a value of $c=1$ when computing portions of the constraint related to directed edges and a value of $c=2$ when computing portions of the constraint related to bidirected edges. As the constraints in Theorem~\ref{thm:constraints} are valid for any $c>0,$ these values were chosen only to make values of $h(\theta)$ under violations of ancestrality, aridity, and bow-freeness to be larger than the tolerance level ($10^{-8}$) of the augmented Lagrangian procedure. As mentioned in Section~\ref{app:greenery}, a scaling factor applied to the hyperbolic tangent function controls the sharpness of approximation of the primal fixing operator. In our experiments we use a scaling factor of $\ln(5000),$ but any sufficiently large value suffices as long as the penalty $h(\theta)$ computed for c-trees is above the tolerance level of the augmented Lagrangian procedure. Finally symmetry of the matrix $\beta$ is enforced by requiring each off-diagonal entry $\beta_{ij}$ and $\beta_{ji}$ are tied to a single free parameter. Positive-definiteness of $\beta$ is guaranteed by construction in the RICF procedure \citep{drton2009computing}.

\subsection*{Choice of Hyperparameters}

We summarize our choice of hyperparameters and justification for these choices in Table~\ref{tab:hyperparams}. Choice of some hyperparameters, such as tolerance levels for RICF and increments in RICF iterations, require little justification as lower tolerance and more iterations can only improve approximation. We set specific values only to cap the run time of our procedure. Choices for most other hyperparameters are based on prior literature.

\subsection*{Converting Estimates of $\theta$ to an ADMG $\G(\theta)$}
The final step of Algorithm~\ref{alg:discovery} returns an ADMG $\G(\theta)$ as follows. We first derive the matrices $\delta$ and $\beta$ from $\theta.$ The structure of the induced ADMG is then given by: $V_i \rightarrow V_j$ exists in $\G$ if $|\delta_{ij}| > \omega$ and $V_i \leftrightarrow V_j$ exists in $\G$ if $|\beta_{ij}| > \omega$ for all $i\not=j.$ Such thresholding is standard in similar continuous optimization structure learning methods, such as \cite{zheng2018dags} and \cite{yu2019dag}, and the threshold can be made arbitrarily small as long as tolerance to $h(\theta)$ is also small. In our experiments we use $\omega=0.05.$

\renewcommand{\arraystretch}{1.5}
\begin{table}[t]
	\centering
	\begin{tabular}{|l|c|p{8cm}|}
		\hline
		\textsc{Hyperparameter} & \textsc{Setting} & \textsc{Justification} \\
		\hline
		Tolerance for $h(\theta)$ & $10^{-8}$ & Numerically close enough to $0$ -- the lower the better. \\
		Max dual ascent iterations & $100$ & Same value as in \cite{zheng2018dags}; convergence is typically achieved within $10$ iterations. \\
		RICF increment $s$ & $1$ & RICF often converges in $10$ steps \citep{drton2009computing, nowzohour2017distributional}. Higher values should be used for larger graphs. \\
		Regularization strength $\lambda$ & $0.05$ & Obtained through manual testing on held-out data derived from Fig.~\ref{fig:examples}(b,c). \\
		Progress rate $r$ & $0.25$ & Same value as in \cite{zheng2018dags}; \cite{yu2019dag}. \\
		Tolerance for RICF & $10^{-4}$ & Numerically close enough to $0$ -- the lower the better. \\
		\hline
	\end{tabular}
	\caption{Hyperparameter settings used for our experiments.}
	\label{tab:hyperparams}
\end{table}

\clearpage

\section{ADDITIONAL RESULTS AND EXPERIMENTS}
\label{app:exprs}

\begin{figure}[t]
	\centering
	\includegraphics[scale=0.35]{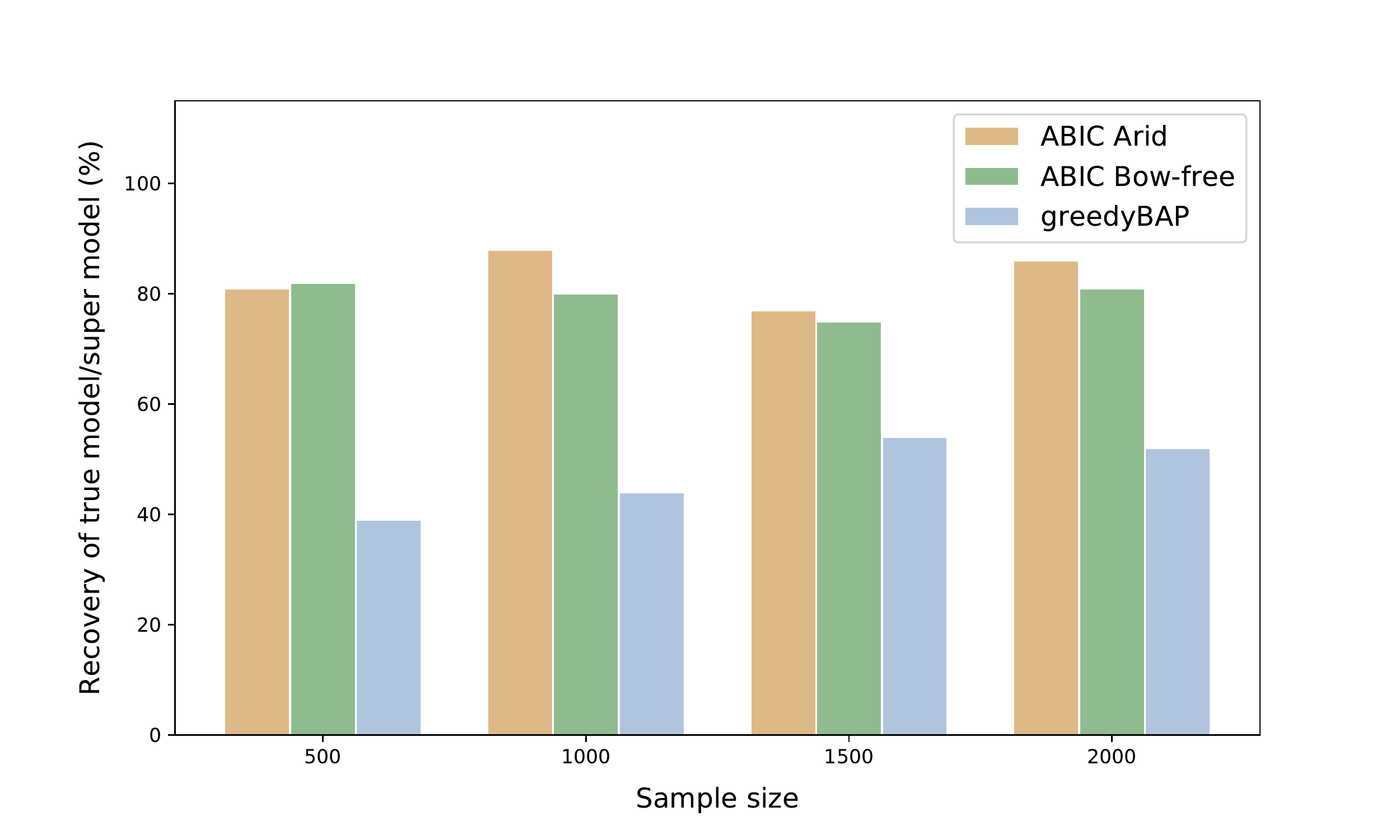}

	\caption{Bar plots showing rate of recovery of the true equivalence class \emph{or} a super model of the true equivalence class of ADMGs with a Verma constraint as a function of sample size. The underlying data is the same as the one used to generate the plots in Fig.~\ref{fig:experiments}.}
	\label{fig:experiments_super}
\end{figure}

\renewcommand{\arraystretch}{1.5}
\begin{table}[t]
	\centering
	\begin{tabular}{|c|c|c|c|} \hline
		$\lambda$     & \textsc{True model} & \textsc{Super model} & \textsc{Wrong model} \\ \hline
		5e-4       & 0.20       & 0.80        & 0.00        \\
		5e-3       & 0.25       & 0.70        & 0.05        \\
		{\ul 5e-2} & {\ul 0.39} & {\ul 0.41}  & {\ul 0.20}  \\
		5e-1       & 0.01       & 0.00        & 0.99        \\
		5e0        & 0.00       & 0.00        & 1.00        \\ \hline
	\end{tabular}
	\caption{Analysis of different settings of $L_0$-regularization parameter $\lambda$ in the ABIC bow-free procedure. We report the fraction of times the procedure recovered the true model (or one that is equivalent to it), a super model of the true model, or an incorrect independence model. The underlying data for the experiment is the same as the one used to generate the bar plots in Fig.~\ref{fig:experiments} for $n=1000$. We use the underlined $\lambda$ = 5e-2 for all experiments.}
	\label{tab:lambda_exprs}
\end{table}

In this section we provide additional results and experiments that were excluded from the main draft due to space constraints.

Fig.~\ref{fig:experiments_super} provides additional insight into the modes of failure for each algorithm used to recover Verma constraints in the experiments corresponding to the bar plots in Fig.~\ref{fig:experiments} of the main draft. It is easy to see from Fig.~\ref{fig:experiments_super} that more often than not, the arid and bow-free ABIC methods yield an equivalent model or a super model of the true ADMG while the greedyBAP method more often returns an incorrect model.

Table~\ref{tab:lambda_exprs} shows the results obtained from the ABIC bow-free procedure for different settings of regularization stength $\lambda.$ Results are shown for the same task as in Fig.~\ref{fig:experiments} of recovering ADMGs with a Verma constraint for sample size $n=1000.$ As expected, for low values of $\lambda,$ the procedure is more likely to return a denser ADMG corresponding to a super model of the true model. As $\lambda$ increases, the procedure recovers the true model more often, and finally for relatively large values of $\lambda$ the procedure almost always returns a sparser ADMG corresponding to an incorrect independence model.

Finally, we present results for 15 variable ADMGs in Table~\ref{tab:results_V15} to supplement the 10 variable experiments in Table~\ref{tab:results} of the main paper. We observe similar trends showing that our method performs favorably in comparison to baselines for recovery of both arid and ancestral ADMGs.

\clearpage

\begin{table*}[t]
	\resizebox{\textwidth}{!}{
		
		\begin{tabular}{l|ll|ll|ll|}
			\cline{2-7}
			& \multicolumn{2}{c|}{\textsc{Skeleton}}  & \multicolumn{2}{c|}{\textsc{Arrowhead}} & \multicolumn{2}{c|}{\textsc{Tail}}    \\ \hline
			\multicolumn{1}{|l|}{Method}        & tpr $\uparrow$ & fdr $\downarrow$ & tpr $\uparrow$& fdr$\downarrow$ & tpr $\uparrow$& fdr $\downarrow$ \\ \hline
			\multicolumn{1}{|l|}{gBAP~\citep{nowzohour2017distributional}} 
			& 0.80       & 0.27       & 0.28       & 0.53       & 0.02       & 0.42       \\
			\multicolumn{1}{|l|}{ABIC (bow-free)}
			& {\bf 0.83} & {\bf 0.15} & {\bf 0.69} & {\bf 0.23} & {\bf 0.26} & {\bf 0.41} \\ \hline
		\end{tabular}
		
		\quad
		
		\begin{tabular}{l|ll|ll|ll|}
			\cline{2-7}
			& \multicolumn{2}{c|}{\textsc{Skeleton}}  & \multicolumn{2}{c|}{\textsc{Arrowhead}} & \multicolumn{2}{c|}{\textsc{Tail}}    \\ \hline
			\multicolumn{1}{|l|}{Method}        & tpr $\uparrow$ & fdr $\downarrow$ & tpr $\uparrow$& fdr$\downarrow$ & tpr $\uparrow$& fdr $\downarrow$ \\ \hline
			\multicolumn{1}{|l|}{FCI~\citep{spirtes2000causation}} 
			& 0.29       & 0.11       & 0.24       & 0.56       & 0.05       & 0.74       \\
			\multicolumn{1}{|l|}{gSPo~\citep{bernstein2020ordering}} 
			& {\bf 0.87}       & 0.23       & 0.41       & 0.62       & 0.31       & 0.88       \\
			\multicolumn{1}{|l|}{ABIC (ancestral)}
			& {0.77} & {\bf 0.09} & {\bf 0.66} & {\bf 0.24} & {\bf 0.62} & {\bf 0.68} \\ \hline
		\end{tabular}
	}
	\caption{Comparison of our method to greedyBAP (left) and FCI (right) for recovering 15 variable arid and ancestral ADMGs respectively. The metrics reported are analogous to Table~\ref{tab:results} in the main text. ($\uparrow$/$\downarrow$ indicates higher/lower is better.)}
	\label{tab:results_V15}
\end{table*}

\bibliography{references}

\makeatletter\@input{mainxx.tex}\makeatother

\bibliography{references}

\end{document}